\documentclass[10pt,twocolumn,letterpaper]{article}
\newcommand{\beginsupplement}{%
\setcounter{section}{0}
\renewcommand{\thesection}{S\arabic{section}}%
\setcounter{table}{0}
\renewcommand{\thetable}{S\arabic{table}}%
\setcounter{figure}{0}
\renewcommand{\thefigure}{S\arabic{figure}}%
}
\usepackage{titling}
\predate{}
\date{}
\postdate{}
\usepackage{cvpr}
\usepackage{times}
\usepackage{epsfig}
\usepackage{graphicx}
\usepackage{amsmath}
\usepackage{amssymb}
\usepackage{subfigure}

\usepackage[pagebackref=true,breaklinks=true,letterpaper=true,colorlinks,bookmarks=false]{hyperref}

\cvprfinalcopy 

\def\cvprPaperID{3468} 

\ifcvprfinal\fi
\begin{document}

\title{Seeing Small Faces from Robust Anchor's Perspective}

\author{Chenchen Zhu \qquad Ran Tao \qquad Khoa Luu \qquad Marios Savvides\\
Carnegie Mellon University\\
5000 Forbes Avenue, Pittsburgh, PA 15213, USA\\
{\tt\small \{chenchez, rant, kluu, marioss\}@andrew.cmu.edu}
}

\maketitle

\begin{abstract}
This paper introduces a novel anchor design to support anchor-based face detection for superior scale-invariant performance, especially on tiny faces. To achieve this, we explicitly address the problem that anchor-based detectors drop performance drastically on faces with tiny sizes, e.g. less than $16 \times 16$ pixels. In this paper, we investigate why this is the case. We discover that current anchor design cannot guarantee high overlaps between tiny faces and anchor boxes, which increases the difficulty of training. The new Expected Max Overlapping (EMO) score is proposed which can theoretically explain the low overlapping issue and inspire several effective strategies of new anchor design leading to higher face overlaps, including anchor stride reduction with new network architectures, extra shifted anchors, and stochastic face shifting. Comprehensive experiments show that our proposed method significantly outperforms the baseline anchor-based detector, while consistently achieving state-of-the-art results on challenging face detection datasets with competitive runtime speed.

\end{abstract}

\section{Introduction}
\label{sec:intro}

\begin{figure}
\centering
\hspace{-6mm}
\subfigure[Recall Rate-Face Scale]{
\includegraphics[width=0.5\columnwidth]{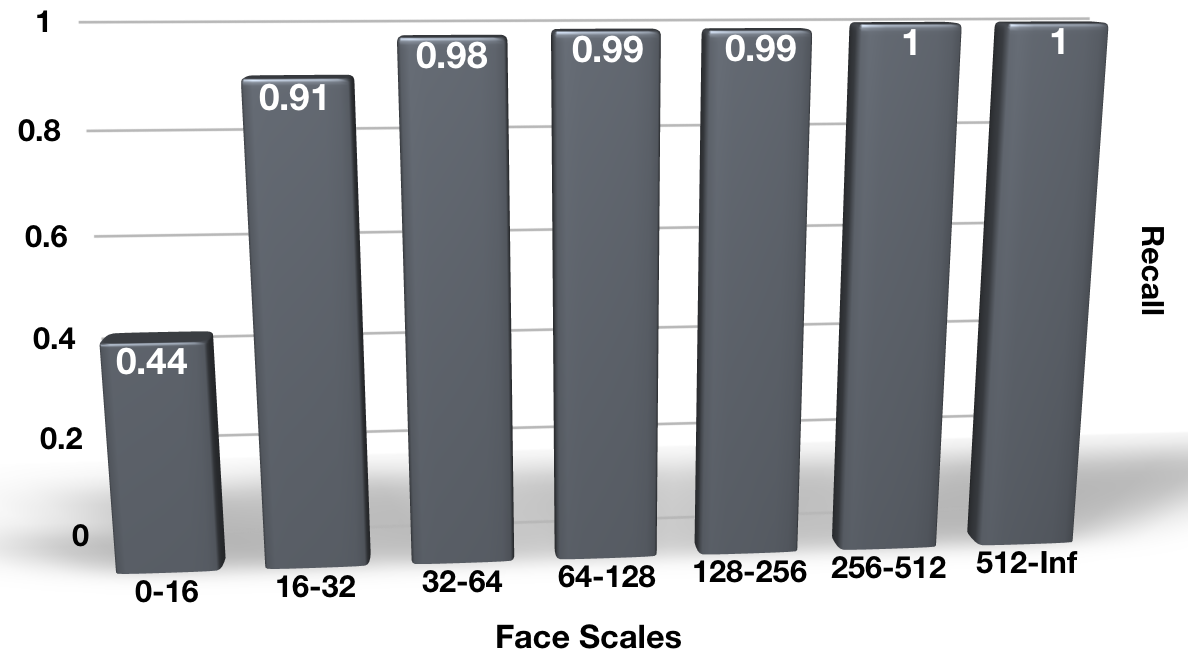}
\label{fig:intro_a}
}
\hspace{-3mm}
\subfigure[Average IoU-Face Scale]{
\includegraphics[width=0.5\columnwidth]{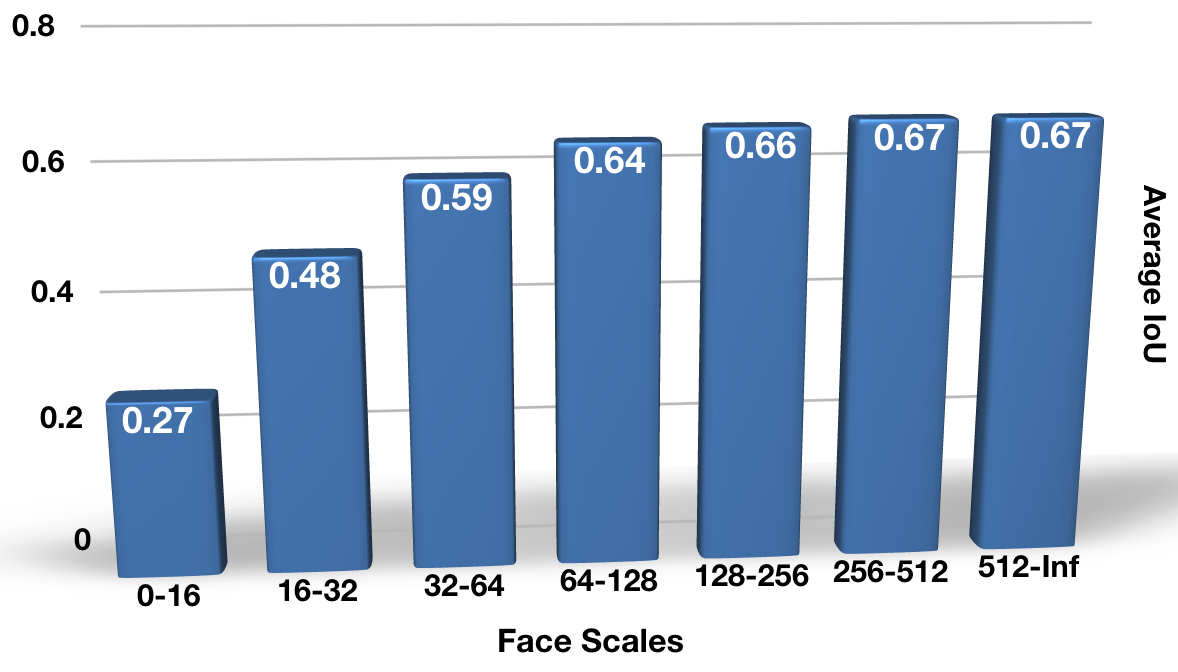}
\label{fig:intro_b}
}
\subfigure[Baseline anchors (yellow) vs. our anchors (red) with higher face IoUs]{
\includegraphics[width=\columnwidth]{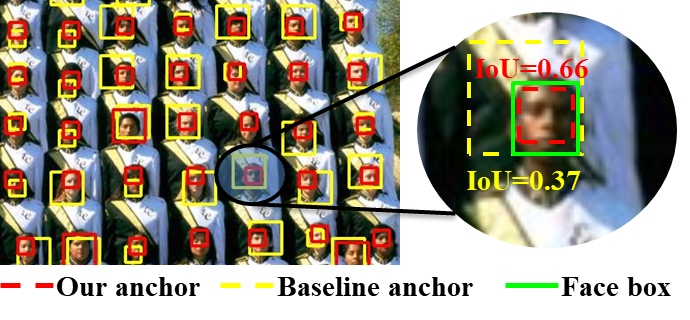}
\label{fig:intro_c}
}
\caption{
Problems of current anchor-based detectors and our solution. \textbf{(a):} A baseline anchor-based detector, trained and evaluated on the Wider Face dataset (see Section \ref{exp:ablation} for details), has significantly lower recall rate at IoU of 0.5 on tiny faces ($16 \times 16$) than larger faces. \textbf{(b):} Maximum IoU with anchors is computed for each face and averaged in each scale group, showing positive correlation with the recall rate across the scales. \textbf{(c):} Visualization of the anchor boxes with highest IoUs for each face. Our anchors have much higher IoU with faces than the baseline anchor. Right side shows an enlarged example. (Best viewed in color)
}
\label{fig:intro}
\end{figure}

Face detection plays an important role in many facial analysis tasks \cite{3dtps, centerloss, sphereface}. Starting from the traditional face detectors with hand-crafted features \cite{viola2001rapid,zhu2012face, Felzenszwalb, yu2013pose,li2013learning}, the modern detectors \cite{faceness, conv3d, ldcf, mtcnn} have been dramatically improved thanks to the robust and discriminative features from deep convolutional neural networks (CNNs) \cite{vgg, he2016resnet}. Current state-of-the-art face detectors are designed using \textit{anchor-based} deep CNNs \cite{cms-rcnn, hu2016tiny, yang2017scaleface, najibi2017ssh, sfd}, inspired by the techniques from popular object detectors \cite{liu2016ssd, ren2015faster, li2016rfcn}.

Anchor-based detectors quantize the continuous space of all possible face bounding boxes on the image plane into the discrete space of a set of pre-defined anchor boxes that serve as references. The Intersection-over-Union (IoU) overlap is used for evaluating the similarity between anchor boxes and face boxes. During training, each face is matched with one or several close anchors. These faces are trained to output high confidence scores and then regress to ground-truth boxes. During inference, faces in a testing image are detected by classifying and regressing anchors.

Although anchor-based detectors have shown success in handling shape and appearance invariance \cite{voc,coco}, their capabilities in handling scale invariance is not satisfactory. 
On the other hand, faces can be captured at any size in images. In addition to heavy occlusion, extreme pose and low illumination, very small faces have become one of the most challenging problems in robust face detection.
Figure \ref{fig:intro_a} shows the face recall rate of a baseline anchor-based detector across different face scales. While big faces (larger than $64 \times 64$ pixels) can be almost 100\% recovered, there is a significant drop in recall for smaller faces, especially those with less than $16 \times 16$ pixels. In other words, after classifying and adjusting anchor boxes, the new boxes with high confidence scores are still not highly overlapped with enough small faces. This suggests that we look at how anchors are overlapped with faces initially before training. For each face we compute its highest IoU with overlapped anchors. Then faces are divided into several scale groups. Within each scale group we compute the averaged highest IoU score, as presented in Figure \ref{fig:intro_b}. It's not surprising to find that average IoUs across face scales are positively correlated with the recall rates. We think anchor boxes with low IoU overlaps with small faces are harder to be adjusted to the ground-truth, resulting in low recall of small faces. 


In this paper, we focus on new anchor design to support anchor-based detectors for better scale-invariance. Our newly proposed anchors have higher IoU overlaps with faces than the baseline anchors, as shown in Figure \ref{fig:intro_c}. Therefore, it is easier for the network to learn how to adjust the new anchors (red boxes) to ground-truth faces than the original anchors (yellow boxes). To achieve this, we look deep into how faces are matched to anchors with various configurations and propose the new Expected Max Overlapping (EMO) score to characterize anchors' ability of achieving high IoU overlaps with faces. Specifically, given a face of known size and a set of anchors, we compute the expected max IoU of the face with the anchors, assuming the face's location is drawn from a 2D distribution of on the image plane. 
The EMO score theoretically explains why larger faces are easier to be highly overlapped by anchors and that densely distributed anchors are more likely to cover faces.

The EMO score enlightens several simple but effective strategies of new anchor design for higher face IoU scores without introducing much complexity to the network. Specifically, we propose to reduce the anchor stride with various network architecture designs. We also propose to add anchors shifted away from the canonical center so that the anchor distribution becomes denser. In addition, we propose to stochastically shift the faces in order to increase the chance of getting higher IoU overlaps. Finally, we propose to match low-overlapped faces with multiple anchors. 

We run extensive ablative experiments to show our proposed method can achieve significant improvement over the baseline anchor-based detector. It also achieves the state-of-the-art results on challenging face detection benchmarks, including Wider Face \cite{yang2016wider}, AFW \cite{afw}, PASCAL Faces \cite{faceevaluation15}, and FDDB \cite{fddbTech}
In summary, the main contributions of this paper are three folds:
\textbf{(1)} Provide an in-depth analysis of the anchor matching mechanism under different conditions with the newly proposed Expected Max Overlap (EMO) score to theoretically characterize anchors' ability of achieving high face IoU scores.
\textbf{(2)} Propose several effective techniques of new anchor design for higher IoU scores especially for tiny faces, including anchor stride reduction with new network architectures, extra shifted anchors, and stochastic face shifting. Demonstrate significant improvement over the baseline anchor-based method.
\textbf{(3)} Achieve state-of-the-art performance on Wider Face, AFW, PASCAL Faces and FDDB with competitive runtime speed.
\section{Related Work}
Face detection is a mature yet challenging computer vision problem. One of the first well performing approaches is the Viola-Jones face detector \cite{viola2001rapid}. Its concepts of boosting and using simple rigid templates have been the basis for different approaches \cite{zhang2010survey, li2013learning}. More recent works on face detection tend to focus on using different models such as a Deformable Parts Model (DPM) \cite{zhu2012face, Felzenszwalb, yu2013pose, chen2014jointcascade, ghiasi2015multireshpm}. Mathias et al. \cite{mathias2014face} were able to show that both DPM models and rigid template detectors have a lot of potential that has not been adequately explored. All of these detectors extract handcrafted features and optimize each component disjointly, which makes the training less optimal.

Thanks to the remarkable achievement of deep convolutional networks on image classification \cite{vgg, he2016resnet} and object detection \cite{ren2015faster, li2016rfcn, liu2016ssd}, deep learning based face detection methods have also gained much performance improvement recently \cite{faceness, conv3d, cms-rcnn, ldcf, mtcnn, yang2017scaleface, hu2016tiny, deepir, najibi2017ssh, sfd}. CNNs trained on large-scale image datasets provide more discriminative features for face detector compared to traditional hand-crafted features. The end-to-end training style promotes better optimization. The performance gap between human and artificial face detectors has been significantly closed.

However, the Wider Face dataset \cite{yang2016wider} pushes the challenge to another level. In addition to heavy occlusion, extreme pose, and strong illumination, the ultra small sizes of faces in crowd images have become one of the most challenging problems in robust face detection. To solve this, CMS-RCNN \cite{cms-rcnn} incorporates body contextual information to help infer the location of faces. HR \cite{hu2016tiny} builds multi-level image pyramids for multi-scale training and testing which finds upscaled tiny faces. 
SFD \cite{sfd} addresses this with scale-equitable framework and new anchor matching strategy. In this paper, we introduce a novel anchor design for finding more tiny faces, leading to state-of-the-art detection performance.

\begin{figure*}
\centering
\subfigure[Anchor setup and distribution]{
\includegraphics[width=0.66\columnwidth]{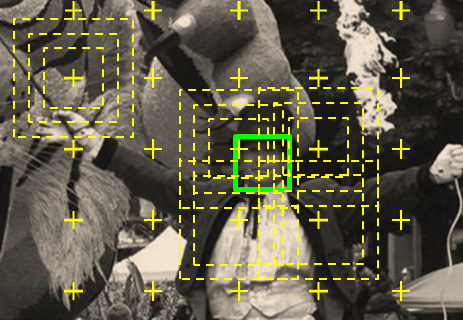}
\label{fig:anchor_matching_a}
}
\subfigure[Anchor matching mechanism]{
\includegraphics[width=0.66\columnwidth]{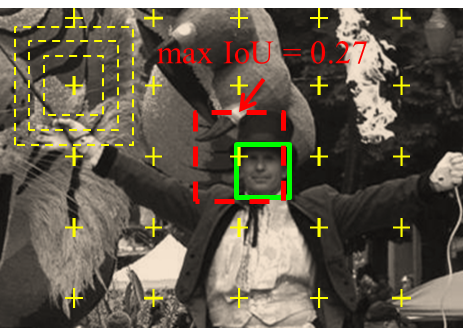}
\label{fig:anchor_matching_b}
}
\subfigure[Computing the EMO score]{
\includegraphics[width=0.66\columnwidth]{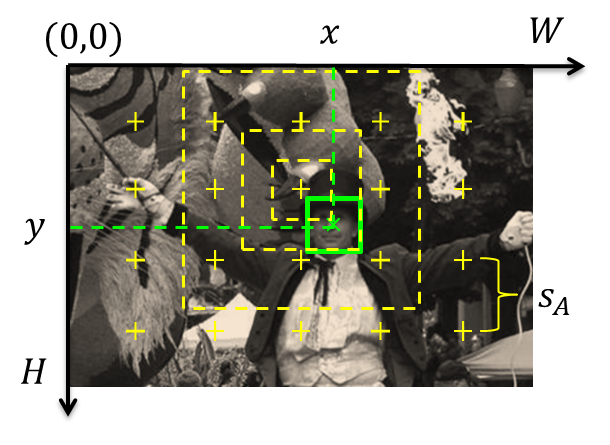}
\label{fig:anchor_matching_c}
}
\caption{\textbf{(a):} Anchors are a set of boxes with different sizes (yellow dashed boxes) tiled regularly (centered on ``+'' crosses) on the image plane. A face (green) overlaps with multiple anchors. \textbf{(b):} A face is matched to an anchor with the max IoU. The matched anchor is highlighted as red dashed box. \textbf{(c):} The EMO score characterizes the anchors' ability of capturing a face by computing the expected max IoU of the face with anchors w.r.t. the distribution of face's location (Best viewed in color).}
\label{fig:anchor_matching}
\end{figure*}

\section{Expected Max Overlapping Scores}
\label{sec:emo}
This section presents the new Expected Max Overlapping (EMO) score to characterize anchors' ability of achieving high face IoU scores. We start with an overview of anchor-based detectors. Then we show the standard anchor setup and the anchor matching mechanism. Finally we derive the EMO by computing the expected max IoU between a face and anchors w.r.t. the distribution of face's location.

\subsection{Overview of Anchor-Based Detector}
\label{subsec:anchor_detector}
Anchor-based detection methods classify and regress anchor boxes to detect objects. Anchors are a set of pre-defined boxes with multiple scales and aspect ratios tiled regularly on the image plane. During training, anchors are matched to the ground-truth boxes based on the IoU overlap. 
An anchor will be assigned to a ground-truth box if a) its IoU with this box is the highest than other anchors, or b) its IoU is higher than a threshold $T_h$. An anchor will be labeled as background if its IoU overlaps with all boxes are lower than a threshold $T_l$.

Anchors are associated with certain feature maps which determine the location and stride of anchors. A feature map is a tensor of size $c \times h \times w$, where $c$ is the number of channels, $h$ and $w$ are the height and the width respectively. It can also be interpreted as $c$-dimensional representations corresponding to $h \cdot w$ sliding-window locations distributed regularly on the image. 
The distance between adjacent locations is the feature stride $s_F$ and decided by $\frac{H}{h}=\frac{W}{w}=s_F$. Anchors take those locations as their centers and use the corresponded representations to compute confidence scores and bounding box regression. So, the anchor stride is equivalent to the feature stride, i.e. $s_A = s_F$.

\subsection{Anchor Setup and Matching}
\label{subsec:anchor_setup}

%
We consider the standard anchor setup as shown in Figure \ref{fig:anchor_matching}. Let $S$ be a pre-defined scale set representing the scales of anchor boxes, and $R$ be a pre-defined ratio set representing the aspect ratios of anchor boxes. Then, the number of different boxes is $|S \times R| = |S||R|$, where $\times$ is the Cartesian product of two sets and $|*|$ is the set's cardinality. For example, anchor boxes with 3 scales and 1 ratio are shown as the yellow dashed rectangles in the top-left corner in Figure \ref{fig:anchor_matching_a}. Let $L$ be the set of regularly distributed locations shown as the yellow crosses ``+'', with the distance between two adjacent locations as anchor stride $s_A$. Then the set of all anchors $A$ is constructed by repeatedly tiling anchor boxes centered on those locations, i.e. $A = S \times R \times L$.

Given a face box $B_f$ shown as the green rectangle, it is matched to an anchor box $B_a$ with the max IoU overlap shown as the dashed red rectangle (Figure \ref{fig:anchor_matching_b}). Then, the max IoU overlap can be computed as in Eq. \eqref{eq:maxbox}.
\begin{equation}
\max_{a \in A}\frac{|B_f \cap B_a|}{|B_f \cup B_a|}
\label{eq:maxbox}
\end{equation}
where $\cap$ and $\cup$ denote the intersection and union of two boxes respectively. 

\subsection{Computing the EMO Score}

\begin{figure}
\centering
\includegraphics[scale=0.6]{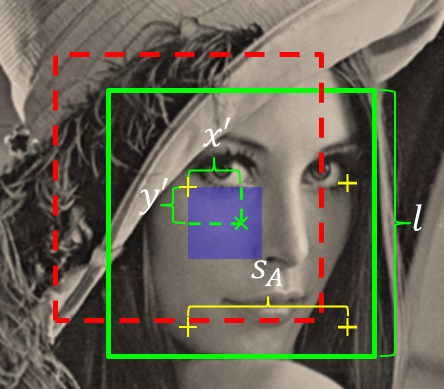}
\caption{The EMO score is a function of face scale $l$ and anchor stride $s_A$ (Best viewed in color).}
\label{fig:emo_cmp}
\end{figure}

A face can randomly appear at any location on the image plane $W \times H$, where $H$ and $W$ are the height and width respectively. In Figure \ref{fig:anchor_matching_c}, we denote the center point of a face as a pair of random variables $(x, y)$, i.e. the green cross ``$\times$''. Let $p(x, y)$ be the probability density function of the location of the face, it then satisfies $\int_0^H \int_0^W p(x, y) dxdy =1$. Plugging in Eq. \eqref{eq:maxbox}, the EMO score can be defined as in Eq. \eqref{eq:emo}.
\begin{equation}
EMO = \int_0^H \int_0^W p(x, y) \max_{a \in A}\frac{|B_f \cap B_a|}{|B_f \cup B_a|} dxdy
\label{eq:emo}
\end{equation}

In practice, we consider the anchor setting according to the Wider Face dataset. We set $S=\{16, 32, 64, 128, 256, 512\}$ to cover the face scale distribution of the dataset, and set $R=\{1\}$ since face boxes are close to squares. Therefore, there are total 6 anchors for each ``+'' location. In addition, we assume each face can randomly appear anywhere on the image plane with equal probability. Thus, $(x, y)$ are drawn from uniform distributions, i.e. $x \sim U(0, W)$ and $y \sim U(0, H)$. 

Since anchors are repeatedly tiled on an image, the overlapping pattern is also periodic w.r.t. the face location. We therefore consider only one period where the face center is enclosed by 4 anchor centers as shown in Figure \ref{fig:emo_cmp}. 
Then, the face will have the highest IoU with the anchor centered on the closest location to the face center. Due to symmetry, we focus on the cases where the face gets matched to the top-left anchor (dashed red box) with the highest IoU and the blue square shows the $\frac{s_A}{2} \times \frac{s_A}{2}$ region where the face center can appear. Face with the center outside that region will be matched to one of the other three anchors. The relative location of the face center to the anchor center is denoted as $(x', y')$, where $x', y'$ are drawn from the distribution $U(0, s_A/2)$. 

Given a $l \times l$ face with the same size as the anchor box, i.e. $l \in S$, it will be matched to the $l \times l$ anchor. So the IoU score between the face and the matched anchor is
\begin{equation}
IoU = \frac{(l-x')(l-y')}{2l^2-(l-x')(l-y')}
\label{eq:iou}
\end{equation}
IoU is a function of the relative location $(x', y')$. Closer distance from face center to anchor center leads to higher IoU.
The EMO score of this face is the expected IoU w.r.t. the distribution of $(x', y')$ derived as in Eq. \eqref{eq:emo_cmp}.
\begin{equation}
EMO=\int_0^{\frac{s_A}{2}} \int_0^{\frac{s_A}{2}} (\frac{2}{s_A})^2 \frac{(l-x')(l-y')}{2l^2-(l-x')(l-y')} dx'dy'
\label{eq:emo_cmp}
\end{equation}

\begin{figure}
\centering
\includegraphics[width=0.7\columnwidth]{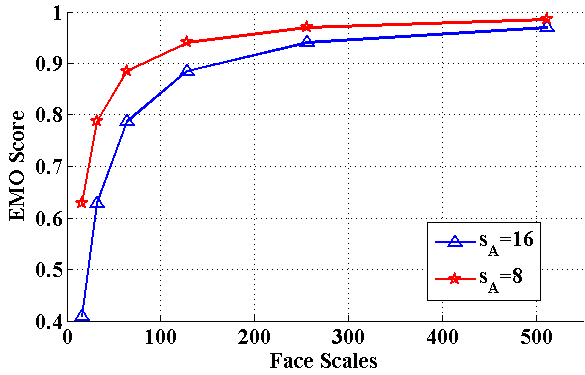}
\caption{The effect of face scale and anchor stride on the EMO score. Small anchor stride is crucial for tiny faces.}
\label{fig:emo_score}
\end{figure}

Figure \ref{fig:emo_score} shows the EMO scores given different face scales and anchor strides. It explains why larger faces tend to have higher IoU overlap with anchors. When the face size is fixed, $s_A$ plays an important role in reaching high EMO scores. Given a face, the smaller $s_A$ is, the higher EMO score achieves. Hence the average max IoU of all faces can be statistically increased.

\section{Strategies of New Anchor Design}
\label{sec:method}
This section introduces our newly designed anchors, with the purpose of finding more tiny faces. We aim at improving the average IoU especially for tiny faces from the view of theoretically improving EMO score, since average IoU scores are correlated with face recall rate.
Based on the analysis in Section \ref{sec:emo}, we propose to increase the average IoU by reducing anchor stride as well as reducing the distance between the face center and the anchor center. 

For anchor stride reduction, we look into new network architectures to change the stride of feature map associated with anchors. Three new architectures are proposed and discussed in Section \ref{subsec:enlarged_feat}. Additionally, we redefined the anchor locations such that the anchor stride can be further reduced in Section \ref{subsec:shifted_anchor}.
Moreover, we propose the face shift jittering method Section \ref{subsec:jittering} which can statistically reduce the distance between the face center and the anchor center, the other important factor to increase the IoU overlap.

With the aforementioned methods, the EMO score can be improved which theoretically guarantee higher average IoU. However, some very tiny faces are still getting low IoU overlaps. We propose a compensation strategy in Section \ref{subsec:compensation} for training which matches very tiny faces to multiple anchors.


\begin{figure*}
\centering
\subfigure[Bilinear upsampling]{
\includegraphics[height=3cm]{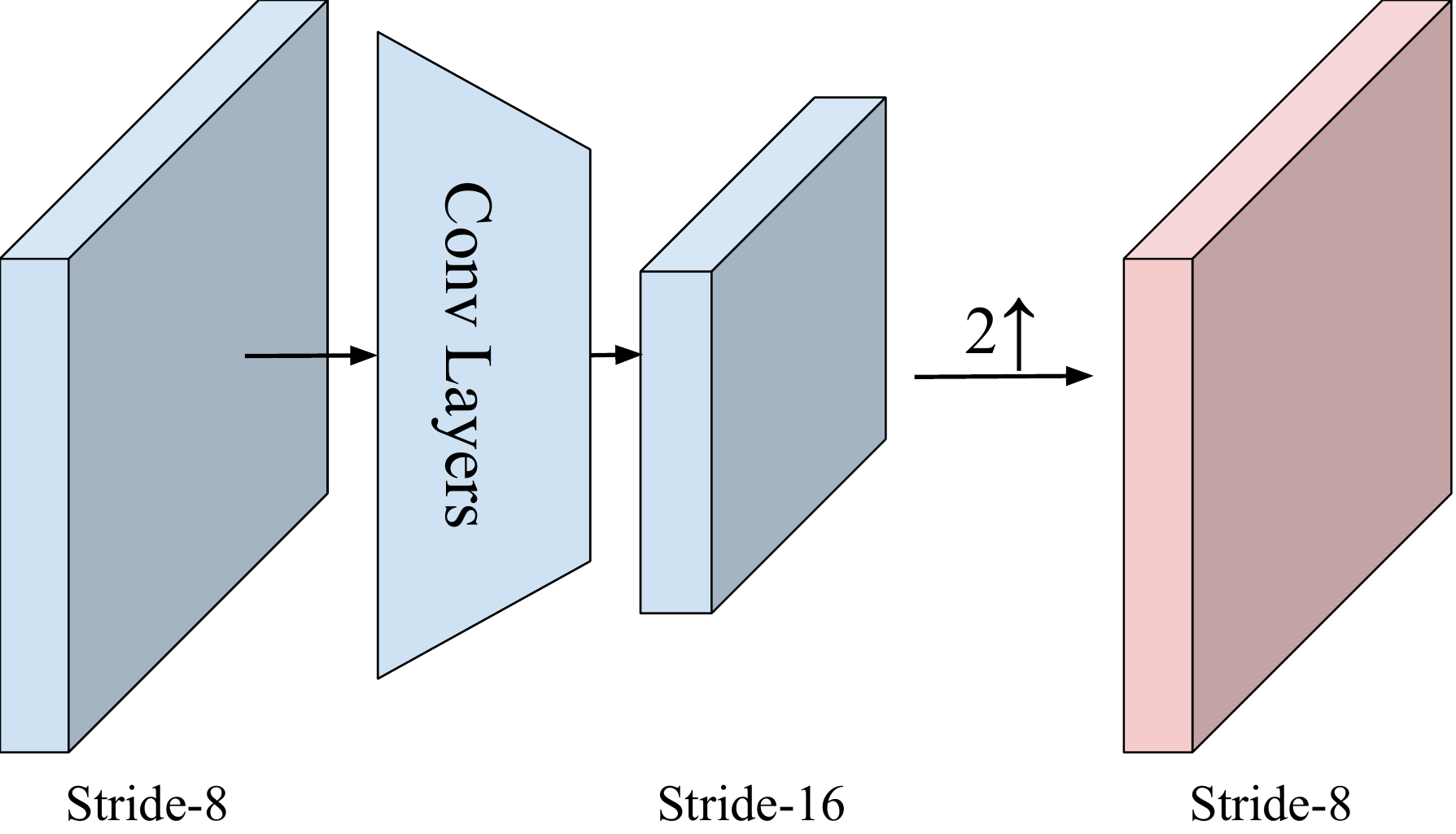}
\label{fig:net_a}}
\subfigure[Bilinear upsampling with skip connection]{
\includegraphics[height=3cm]{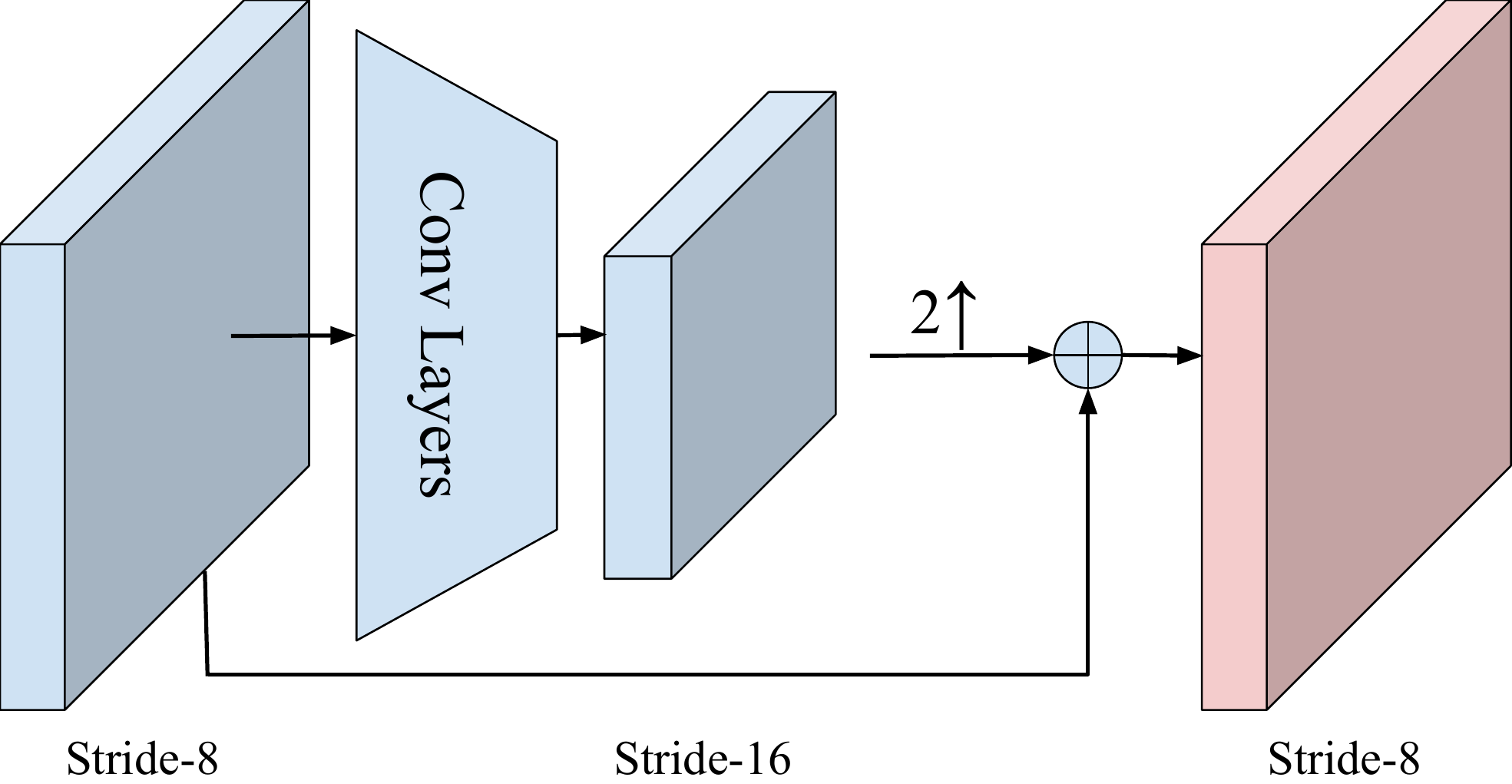}
\label{fig:net_b}}
\subfigure[Dilated convolution]{
\includegraphics[height=3cm]{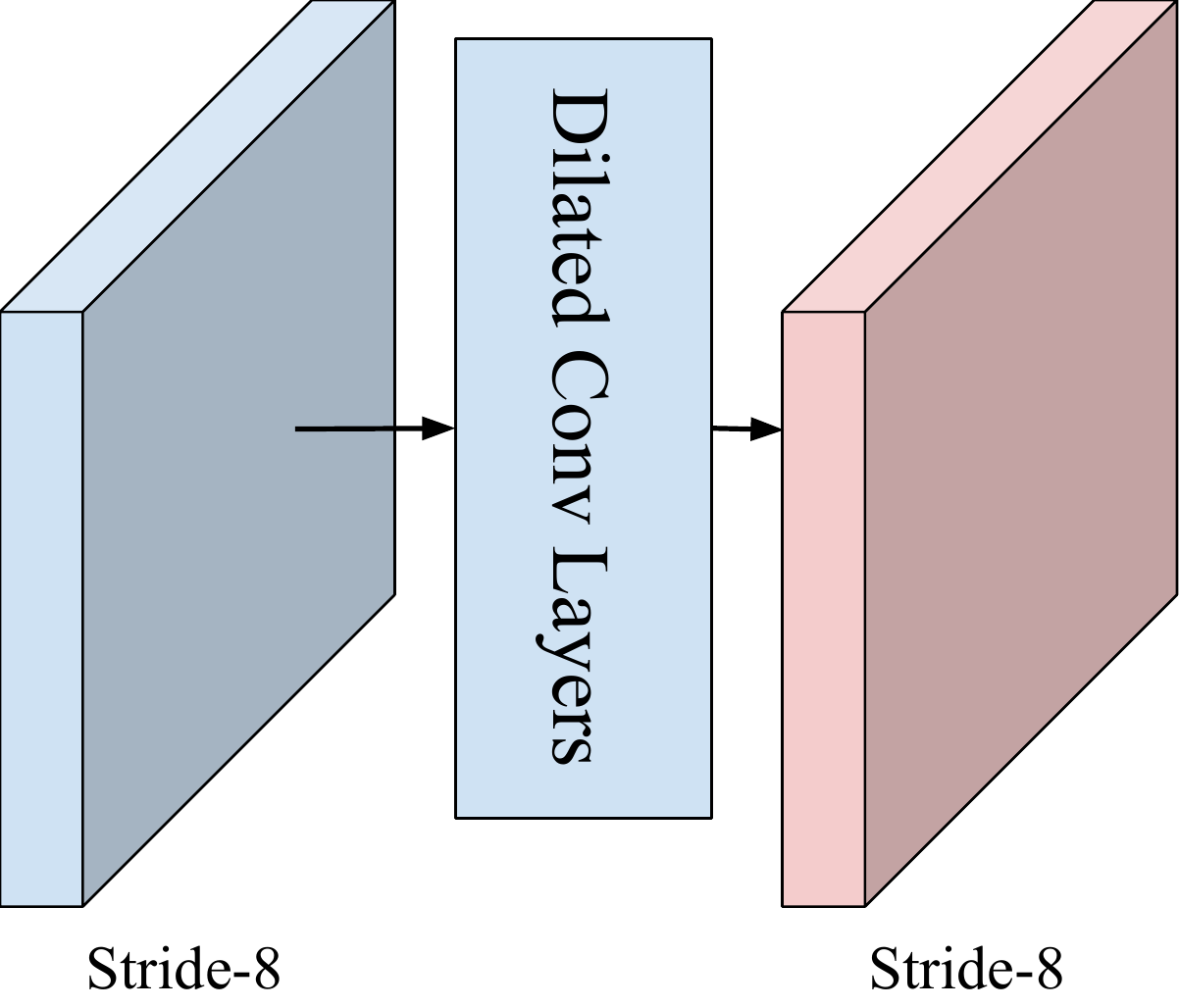}
\label{fig:net_c}}
\caption{Three types of network architecture for reducing the anchor stride by enlarging the feature map (red).}
\label{fig:net}
\end{figure*}

\subsection{Stride Reduction with Enlarged Feature Maps}
\label{subsec:enlarged_feat}

As discussed in Section \ref{subsec:anchor_detector}, anchor stride equals to feature stride in current anchor-based detectors. Therefore, one way to increase the EMO scores is to \textit{reduce the anchor stride} by enlarging the feature map. This section presents three different architectures that double the height and width of the feature maps as illustrated in Figure \ref{fig:net}. 

Bilinear upsampling upscales the feature map by a factor of 2 as shown in Figure \ref{fig:net_a}. In this network, a deconvolution layer is attached to the feature map and its filters are initialized to have weights of a bilinear upsampler. During training, the filters are updated to adapt to the data.

Figure \ref{fig:net_b} shows the upscaled feature map augmented with the features from shallower large feature map by skip connection. The intuition in this design is to combine high level features for semantical information and low level features for localization precision \cite{lin2016fpn}. In the actual networks, the low-level and high-level feature maps have different numbers of channels. Thus, two $1 \times 1$ convolution layers are first added to reduce the number of channels to the same size. Then, after the element-wise addition, another $3 \times 3$ convolution layer is appended to the final feature map for detection (not shown in Figure \ref{fig:net_b}).

The architectures in Figures \ref{fig:net_a} and \ref{fig:net_b} introduce additional parameters to the networks when enlarging the feature maps, hence increasing the model complexity. However, the same goal can be achieved without additional parameters by using dilated convolutions \cite{dilation} as shown in Figure \ref{fig:net_c}. Specifically, we take out the stride-2 operation (either pooling or convolution) right after the shallower large map and dilate the filters in all the following convolution layers. Note that $1 \times 1$ filters are not required to be dilated. In addition to not having any additional parameters, dilated convolution also preserve the size of receptive fields of the filters. 

With halved anchor stride, the average IoU of tiny faces increases by a large margin as shown in Figure \ref{fig:iou_cmp}, compared to the original one. In addition, we show in Section \ref{exp:ablation} the performance comparison of the three architectures.

\subsection{Extra Shifted Anchors}
\label{subsec:shifted_anchor}

Reducing anchor strides by enlarging feature maps doesn't change the condition that $s_A = s_F$.
This section presents a new approach such that $s_A < s_F$. We further reduce $s_A$ by adding \textit{extra supportive anchors} not centered on the sliding window locations, i.e. \textit{shifted anchors}. This strategy can help to increase the EMO scores without changing the resolution of feature maps. These shifted anchors share the same feature representation with the anchors in the centers.

\begin{figure}
\centering
\subfigure[$s_A=s_F$]{
\includegraphics[width=0.33\columnwidth]{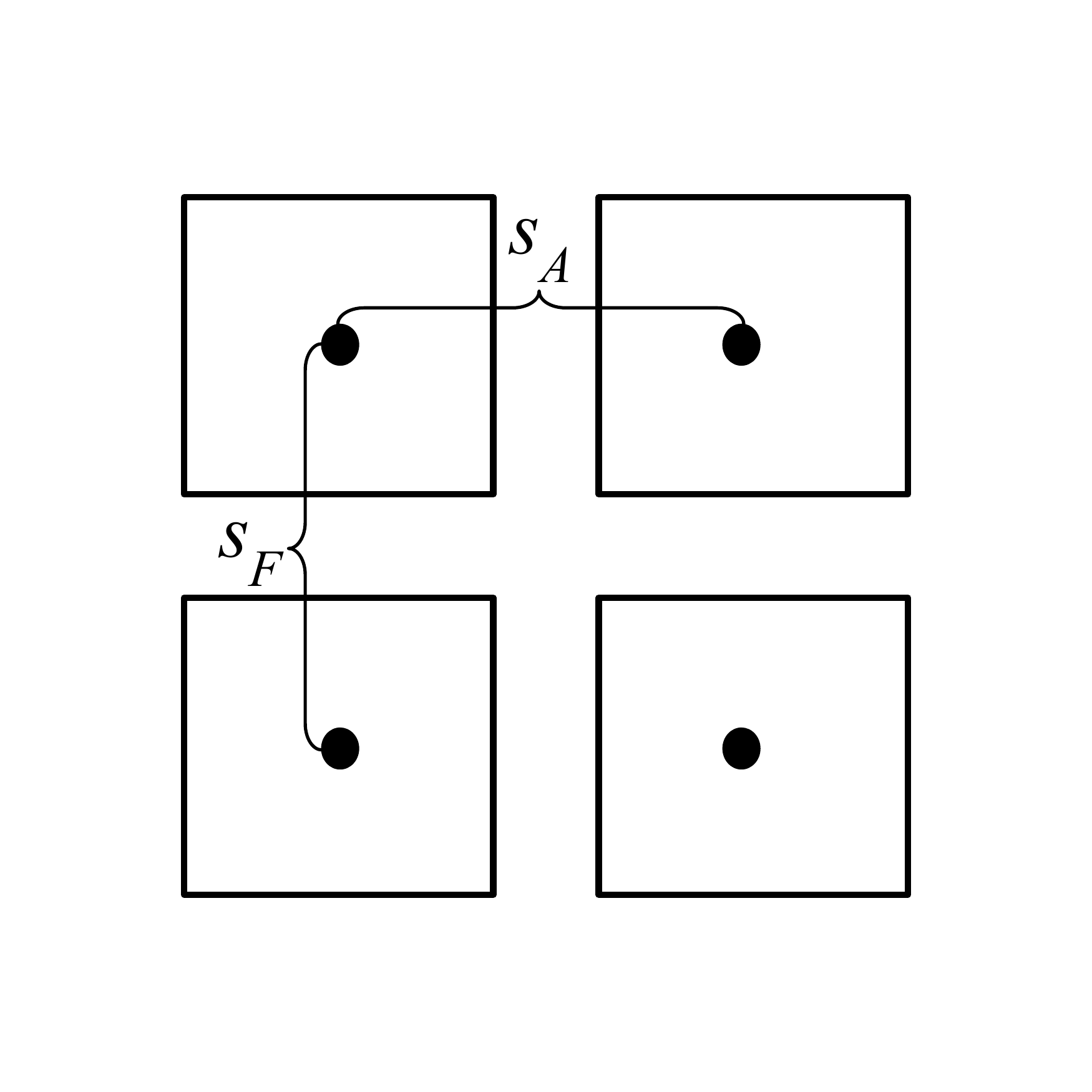}
\label{fig:shift_anchor_a}}
\hspace{-5mm}
\subfigure[$s_A=s_F/\sqrt{2}$]{
\includegraphics[width=0.33\columnwidth]{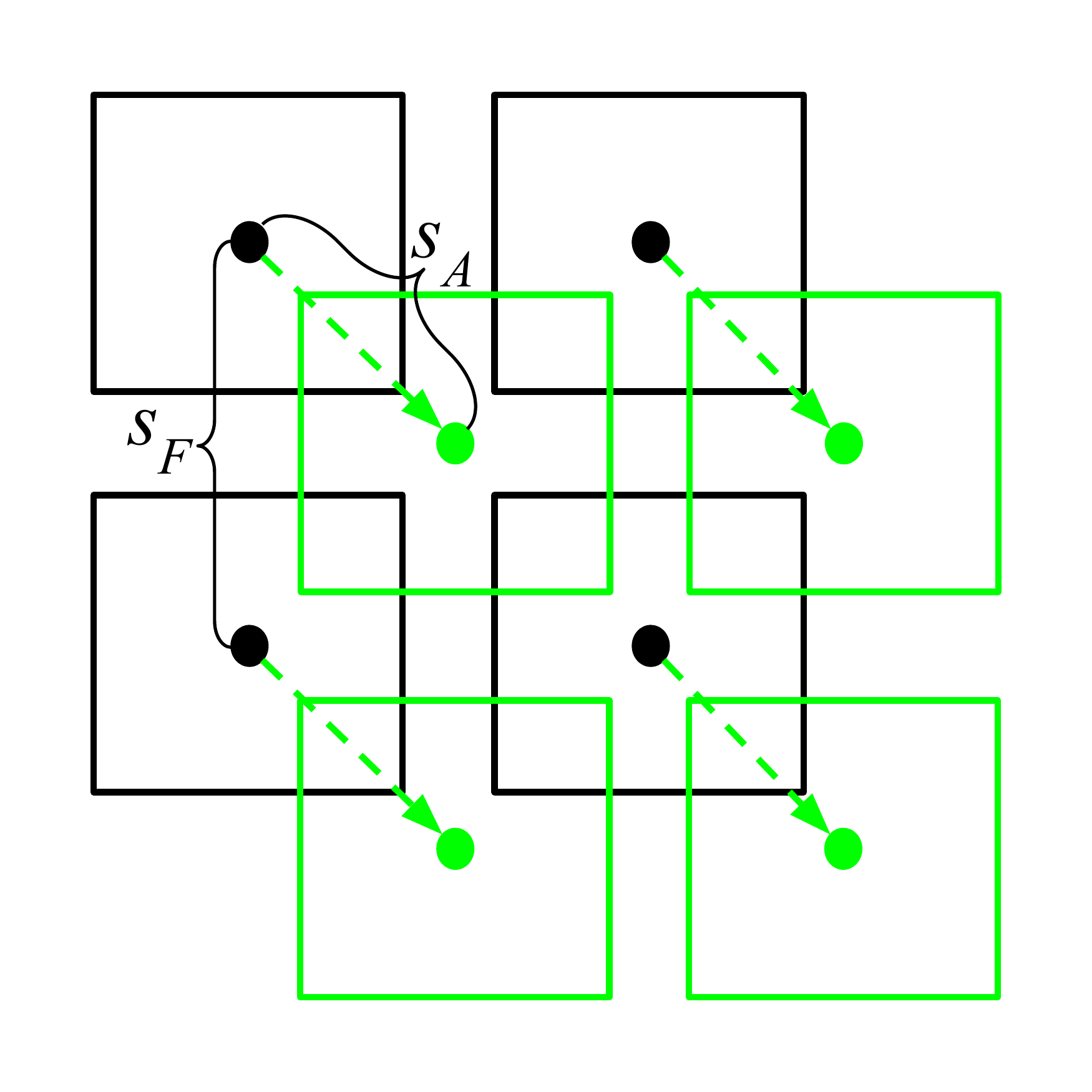}
\label{fig:shift_anchor_b}}
\hspace{-5mm}
\subfigure[$s_A=s_F/2$]{
\includegraphics[width=0.33\columnwidth]{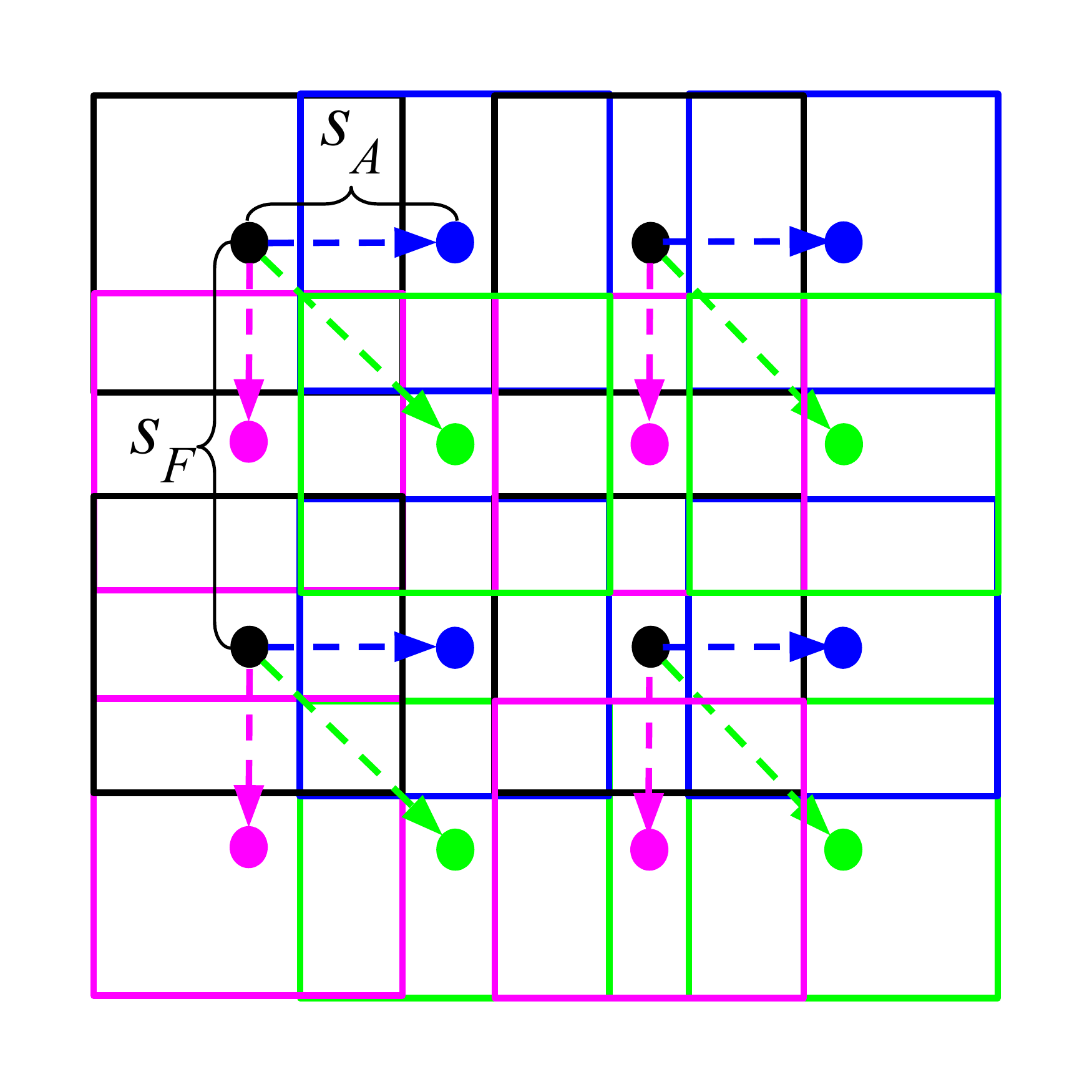}
\label{fig:shift_anchor_c}}
\caption{Anchor stride reduction by adding shifted anchors.}
\label{fig:shift_anchor}
\end{figure}

\begin{figure}[t]
\centering
\includegraphics[width=0.8\columnwidth]{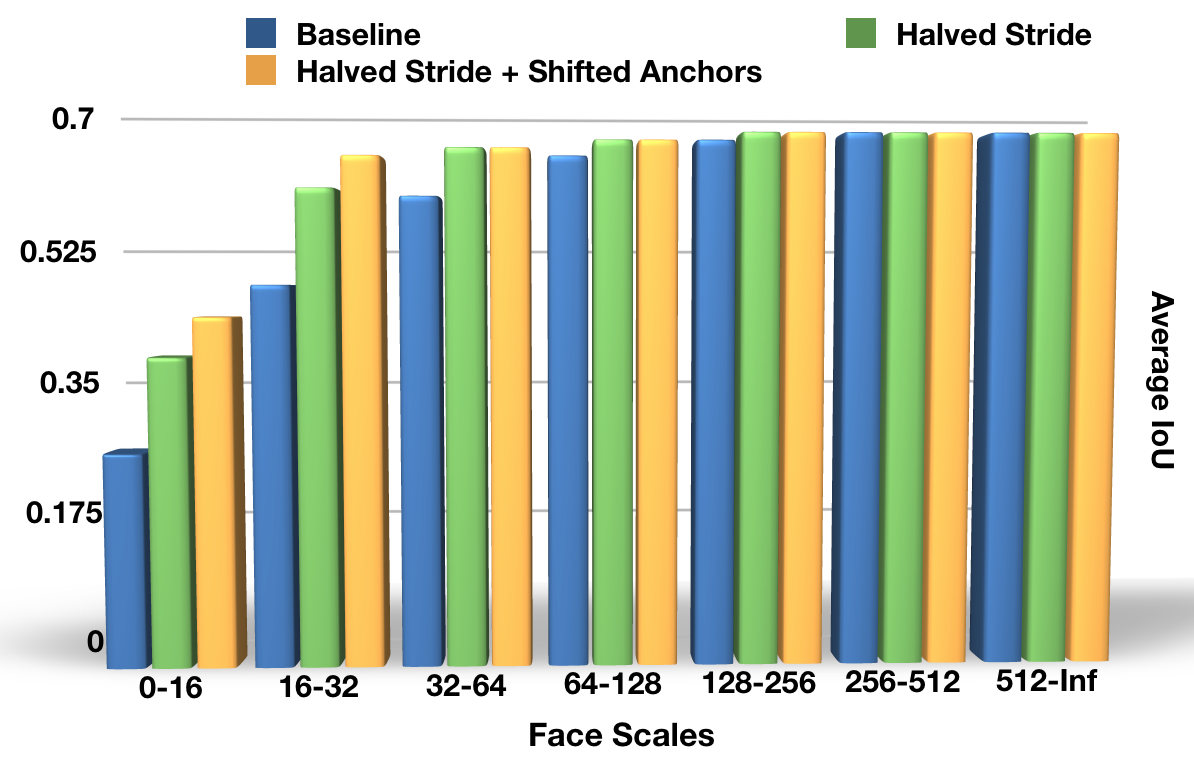}
\caption{Comparison of the average IoU between original and our anchor design. With our proposed stride reduction techniques, the average IoU is significantly improved for small faces.}
\label{fig:iou_cmp}
\end{figure}

Specifically, given a feature map with stride $s_F$, the distance between two adjacent sliding window locations is $s_F$, and labeled by black dots in Figure \ref{fig:shift_anchor}. In Figure \ref{fig:shift_anchor_a}, each location has a single anchor (black) centered on it, giving the anchor stride of $s_A=s_F$. When extra supportive (green) anchors are added to the bottom-right of the center for all locations, the anchor stride can be reduced to $s_A=s_F/\sqrt{2}$ (Figure \ref{fig:shift_anchor_b}). In addition, two other supportive anchors (blue and magenta) can be sequentially added to further reduce the anchor stride to $s_A=s_F/2$ (Figure \ref{fig:shift_anchor_c}). Note that no matter how many anchors are added, all anchors are regularly tiled in the image plane. Indeed, we only need to add small shifted anchors since large anchors already guarantee high average IoUs, which saves the computational time. For example, three shifted anchors of the smallest size ($16 \times 16$) are added on top of enlarged feature maps (Section \ref{subsec:enlarged_feat}) to show further improvement of the average IoUs in Figure \ref{fig:iou_cmp}.


\subsection{Face Shift Jittering}
\label{subsec:jittering}

When computing the EMO score for each face, the center of the face is assumed to be drawn from a 2D uniform distribution. However in the real world datasets, each face has a fixed location. Some of them are closed to the anchor centers so they are more likely to have high IoU overlaps with anchors. While some others far from anchor centers will always get low IoU scores. In order to increase the probability for those faces to get high IoU overlap with anchors, they are randomly shifted in each iteration during training. 

Specifically, the image is shifted by a random offset $(\delta_x, \delta_y)$ in every iteration. $\delta_x$ and $\delta_y$ are the pixels where the image is shifted to right and down respectively, so that locations of all faces in that image are added by $(\delta_x, \delta_y)$. The offset is sampled from a discrete uniform distribution, i.e. $\delta_x, \delta_y \in \{0, 1, ..., s_A/2-1\}$. We use discrete uniform distribution of offsets to approach the continuous uniform distribution of face locations. We set the maximum offset to be $s_A/2-1$ because the period of overlap pattern is $s_A/2$.

\subsection{Hard Face Compensation}
\label{subsec:compensation}
As shown in Figure \ref{fig:iou_cmp}, even with halved feature strides and shifted small anchors, very tiny faces still have lower average IoU than bigger faces. It is because face scales and locations are continuous whereas anchor scales and locations are discrete. Therefore, there are still some faces whose scales or locations are far away from the anchor. These faces are hard to be matched to anchors.

We propose a compensation strategy of anchor matching to assign hard faces to multiple anchors. Specifically, we first label anchors as positive if their overlapping scores with faces are higher than a threshold $T_h$, same as the current anchor matching strategy. Then faces whose highest overlapping scores are below $T_h$ are the hard faces. For hard faces the top $N$ anchors with highest overlap with them are labeled as positive. We find the optimal $N$  empirically as described in Section \ref{exp:ablation}.

\section{Experiments}
In this section, we first show the effectiveness of our proposed strategies with comprehensive ablative experiments. Then with the final optimal model, our approach achieves state-of-the-art results on face detection benchmarks. The computational time is finally presented.

\subsection{Ablation Study}
\label{exp:ablation}
The Wider Face dataset \cite{yang2016wider} is used in this ablation study. This dataset has 32,203 images with 393,703 labeled faces with a high degree of variability in scales, occlusions and poses. The images are split into training (40\%), validation (10\%) and testing (50\%) set. Faces in this dataset are classified into Easy, Medium, and Hard subsets according to the difficulties of detection. The hard subset includes a lot of tiny faces. 
All networks are trained on the training set and evaluated on the validation set. Average Precision (AP) score is used as the evaluation metric. Note that we train and evaluate on the \textit{original} images \textit{without} rescaling since we want to test the detector's ability of finding real tiny faces instead of upscaled tiny faces.

\textbf{Baseline setup} We build an anchor-based detector with ResNet-101 \cite{he2016resnet} inspired by the R-FCN \cite{li2016rfcn} as our baseline detector. It differs from the original R-FCN in the following aspects. Firstly we set 6 anchors whose scales are from the set $\{16, 32, 64, 128, 256, 512\}$ and all anchors' aspect ratios is 1. This setting matches with the face boxes in the Wider Face dataset. Secondly, ``res5'' is used to generate region proposals instead of ``res4''.
Thirdly, the threshold of IoU for positive anchors is changed to 0.5. All the other settings follow the original \cite{li2016rfcn}. The baseline detector is trained on the Wider Face training set for 8 epochs. The initial learning rate is set to 0.001 and decreases to 0.0001 after 5 epochs. During training we applied online hard example mining \cite{ohem} with the ratio between positives and negatives as 1:3. The detector is implemented in the MXNet framework \cite{mxnet} based on the open source code from \cite{deformable}.

\textbf{The effect of enlarged feature map}
As discussed in Section \ref{subsec:enlarged_feat}, enlarged feature map reduces the anchor stride so that it helps to increase the EMO scores and the average IoU, especially for tiny faces. To better understand its effect on the final detection performance, we apply three architectures in Figure \ref{fig:net} to the ResNet backbone architecture of the baseline detector. The corresponding stride-8 feature and stride-16 feature in the backbone architecture are ``res3'' and ``res5'' respectively. For bilinear upsampling (denoted as \textbf{BU}), ``res5'' is appended with a stride-2 deconvolution layer with filters initialized by a bilinear upsampler. For bilinear upsampling with skip connection (denoted as \textbf{BUS}), all the additional convolution layers have 512 output channels. For dilated convolution (denoted as \textbf{DC}), all the $3 \times $3 filters in ``res4'' and ``res5'' are dilated. Evaluation results are presented as ``BU'', ``BUS'' and ``DC'' in Table \ref{tab:ablation}. Compared to the baseline, the enlarged feature map provides significant improvements on the Hard subset (rising by 7.4\% at most), which mainly consists of tiny faces. Among the three architectures, BUS is better than BU at finding tiny faces, but has more false positives, because BUS uses features from the shallower ``res3'' layer that is less discriminative. DC achieves the best performance on all three subsets without introducing extra parameters. Thus, we fix the architecture to DC in the following experiments.

\textbf{The effect of additional shifted anchors}
Adding shifted anchors can further reduce the anchor stride without changing the resolution of the feature map. In Figure \ref{fig:iou_cmp} we can see with halved anchor stride, faces larger than $32 \times 32$ pixels already have the same average IoU overlap. So we mainly focus on adding anchors with scales of 16 and 32. As shown in Figure \ref{fig:shift_anchor}, there are two settings of shifted anchors, i.e. added by one to reduce the stride by $1/\sqrt{2}$ or added by three to reduce the stride by $1/2$. We denote the shifted anchor setting as $s \times n$, where $s \in \{16, 32\}$ is the scale of anchor and $n \in \{1, 3\}$ is the number of additional anchors. Noted that we always start with adding anchors of scale 16 since larger anchors cannot affect the average IoU of smaller faces. Hence there are total 5 combinations as presented in the second row section in Table \ref{tab:ablation}. 
It turns out that adding shifted anchors can improve the AP score on the Hard set. However, there is a trade-off between number of additional anchors and the detection accuracy. More anchors do not always lead to higher average precision, because each anchor is associated with a set of parameters in the network to predict the confidence score and box offsets. As the number of anchors increases, each anchor is matched to fewer faces and the associated parameters are trained from fewer examples. In the end, we find adding just 3 anchors of scale 16 gives the best performance. 

\textbf{The effect of face shift jittering} 
Next we look into the effect of randomly shifting the faces for each iteration during training, denoted as SJ for shift jittering. It is applied to both the model with DC and the model with DC and 3 shifted anchors of scale 16. Experiments show that shift jittering can further improve the AP scores of Hard faces. In Table \ref{tab:ablation}, the AP rises by 0.9\% on the model with DC (+DC vs. +DC+SJ) and by 0.3\% on the model with DC and 3 shifted anchors of scale 16 (+DC+16x3 vs. +DC+16x3+SJ). 

\textbf{The effect of hard face compensation} The hard face compensation completes our final model. It is denoted as HC($N$) where $N$ is the number of anchors to which the hard faces are assigned. We find $N=5$ is a proper choice since smaller $N$ leads to lower recall rate and larger $N$ results in more false positives. To this end, we denote ``+DC+16x3+SJ+HC(5)'' as our ``Final'' detector.


\textbf{The effect of testing size} The size of the testing images has significant impact on the face detection precision, especially for tiny faces. 
Therefore we evaluate our final model on the Hard set of Wider Face validation set with different image sizes, comparing with the state-of-the-art SSH face detector \cite{najibi2017ssh}. We show in Table \ref{tab:scale} that our detector trained with single scale outperforms the SSH detector trained with multiple scales at every testing size. Note that at the maximum input size of 600x1000, our detector outperforms SSH by 7.1\%, showing the effectiveness of our proposed techniques for detecting tiny faces.

\begin{table}
\centering
\caption{Ablative study of each components in our proposed method on Wider Face validation set. Network architectures: \textbf{BU} - bilinear upsampling; \textbf{BUS} - bilinear upsampling with skip connection; \textbf{DC} - dilated convolution. Extra shifted anchors: $s \times n$ - adding $n$ shifted $s$-by-$s$ anchors. \textbf{SJ} - Face Shift Jittering. \textbf{HC($N$)} - assigning each hard face to top $N$ anchors.}
\begin{tabular}{c|c c c}
\hline \hline
 & Easy & Medium & Hard \\ \hline
Baseline & 0.934 & 0.895 & 0.705 \\ 
+BU & 0.933 & 0.901 & 0.710 \\
+BUS & 0.926 & 0.899 & 0.778 \\
+DC & 0.936 & 0.911 & 0.779 \\ \hline
+DC+16x1 & 0.934 & 0.908 & 0.781 \\
+DC+16x1\&32x1 & 0.936 & 0.910 & 0.782 \\
+DC+16x3 & 0.938 & 0.912 & 0.786 \\
+DC+16x3\&32x1 & 0.937 & 0.909 & 0.781 \\
+DC+16x3\&32x3 & 0.938 & 0.913 & 0.779 \\ \hline
+DC+SJ & 0.939 & 0.910 & 0.788 \\
+DC+16x3+SJ & 0.940 & 0.914 & 0.789 \\ \hline
+DC+16x3+SJ+HC(3) & 0.938 & 0.912 & 0.793 \\
+DC+16x3+SJ+HC(5) (Final) & \textbf{0.940} & \textbf{0.914} & \textbf{0.795} \\
+DC+16x3+SJ+HC(7) & 0.936 & 0.912 & 0.791 \\ \hline \hline
Baseline+Pyramid & 0.943 & 0.927 & 0.840 \\
Final+Pyramid & \textbf{0.949} & \textbf{0.933} & \textbf{0.861} \\ \hline
\end{tabular}
\label{tab:ablation}
\end{table}

\begin{table}
\centering
\caption{The effect of testing size on the average precision (AP) of Hard faces in Wider Face validation set.}
\setlength\tabcolsep{3pt}
\begin{tabular}{c|c c c c}
\hline \hline
Max size & 600x1000 & 800x1200 & 1200x1600 & 1400x1800 \\ \hline
SSH \cite{najibi2017ssh} & 0.686 & 0.784 & 0.814 & 0.810 \\
Ours & \textbf{0.757} & \textbf{0.817} & \textbf{0.838} & \textbf{0.835} \\ \hline
\end{tabular}
\label{tab:scale}
\end{table}

\textbf{The effect of image pyramid} Image pyramid for multi-scale training and testing helps improving the detection performance, as shown by ``Baseline+Pyramid'' in Table \ref{tab:ablation}. By applying our strategies we observe another improvement (``Final+Pyramid''). We follow the same way in \cite{najibi2017ssh} to build the pyramid.

\subsection{Evaluation on Common Benchmarks}
\label{exp:benchmarks}

\begin{figure*}[t]
\centering
\hspace{-5mm}
\subfigure[Easy] {
\includegraphics[width=0.7\columnwidth]{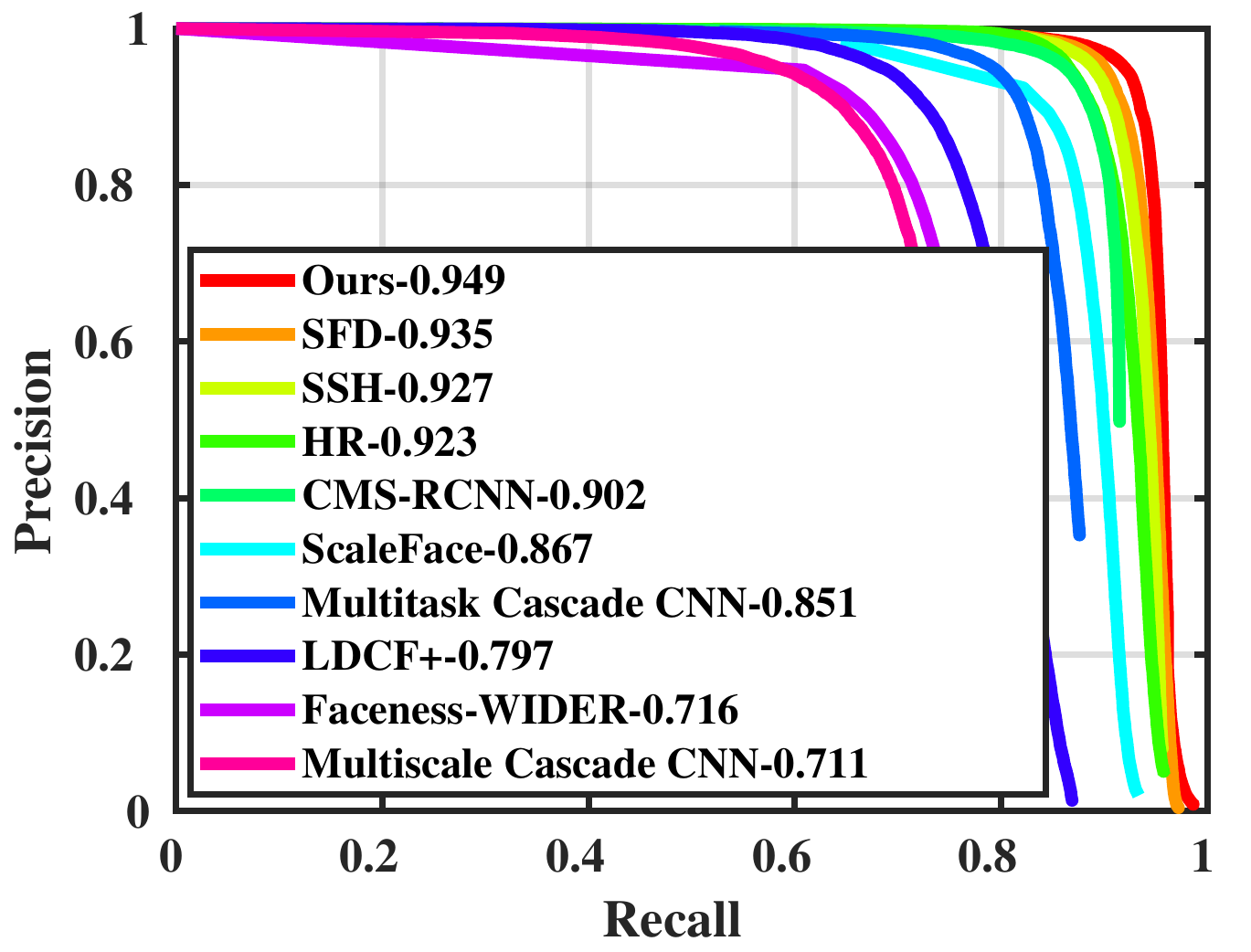}
}
\hspace{-5mm}
\subfigure[Medium] {
\includegraphics[width=0.7\columnwidth]{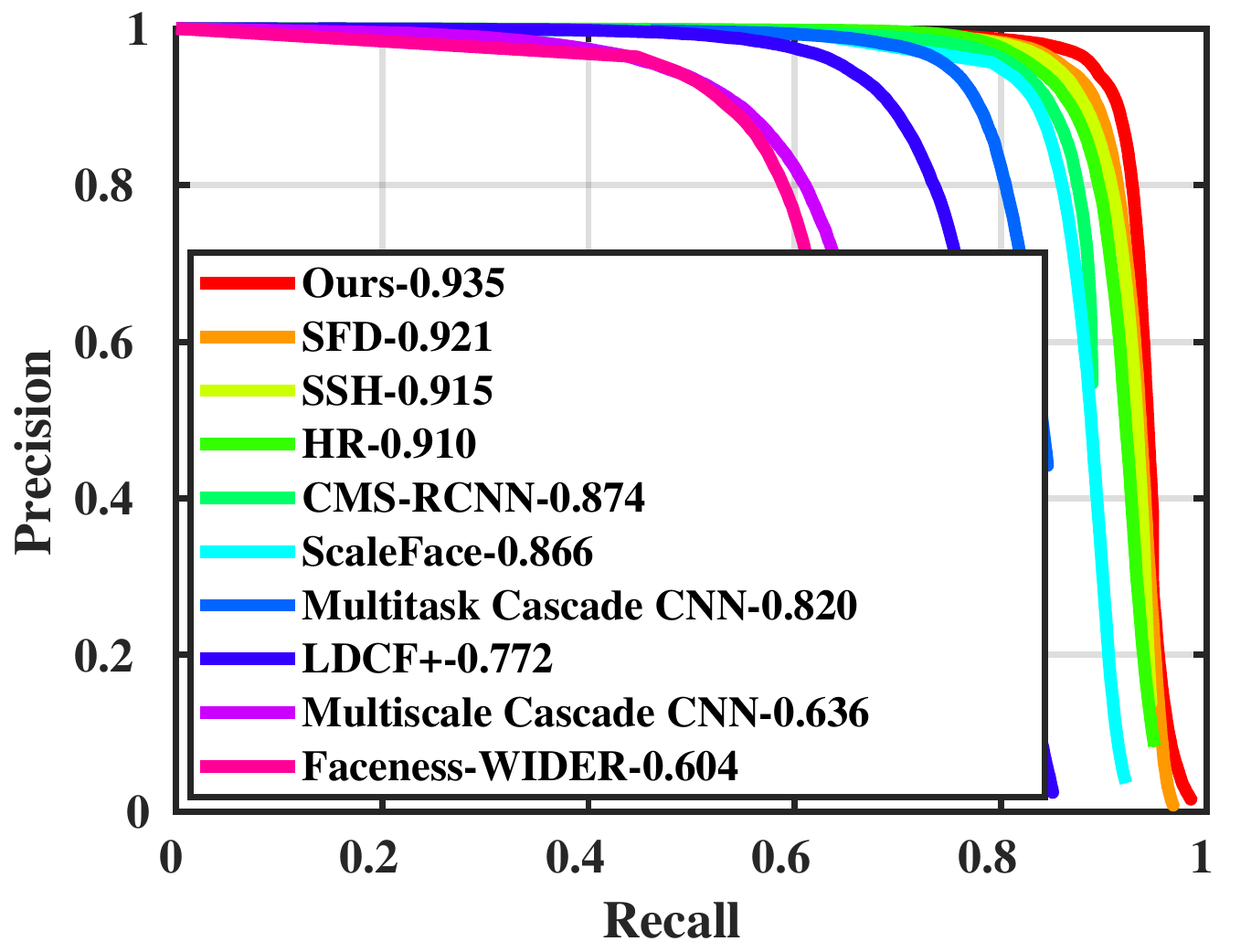}
}
\hspace{-5mm}
\subfigure[Hard] {
\includegraphics[width=0.7\columnwidth]{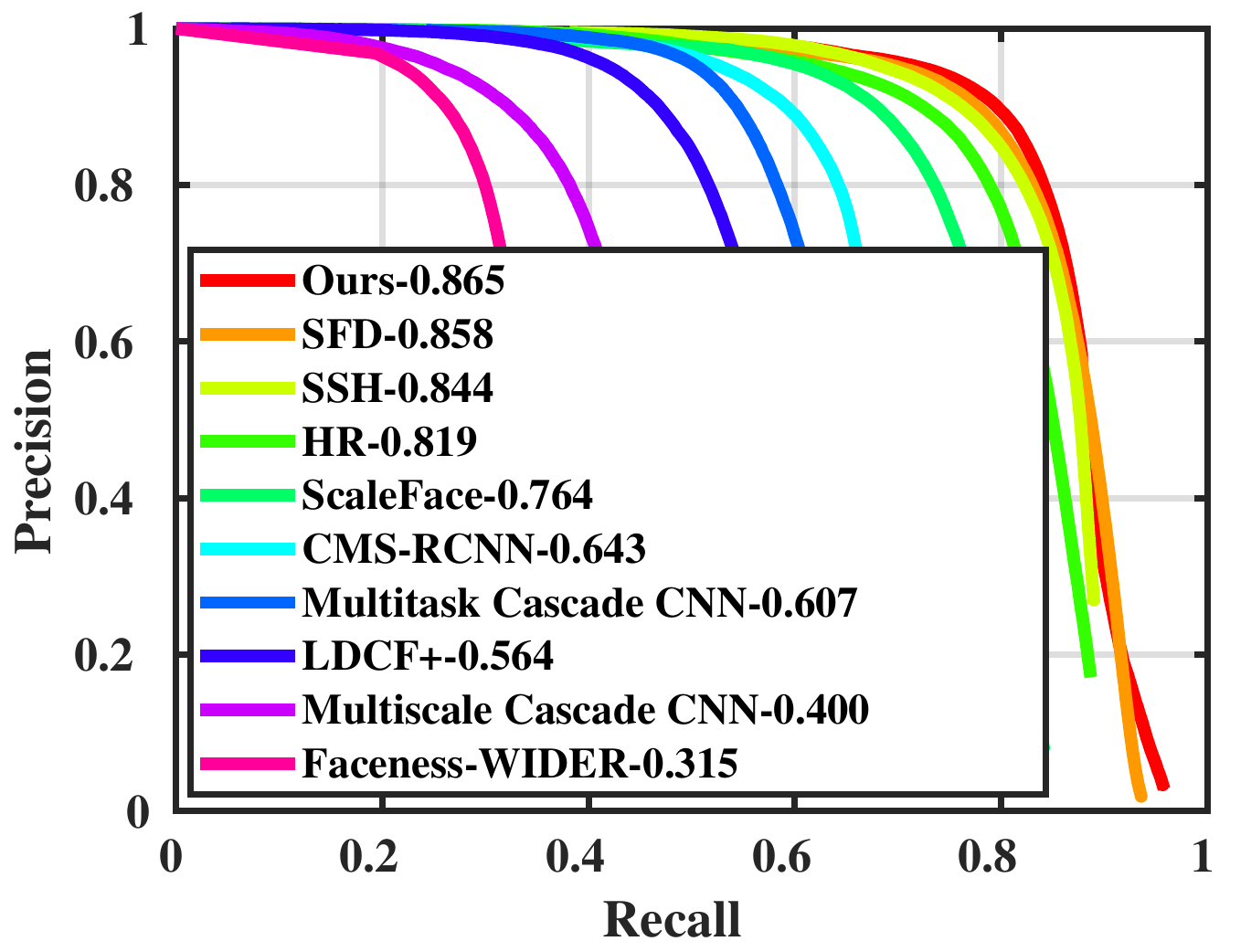}
}
\hspace{-5mm}
\caption{Precision and recall curves on Wider Face testing set divided into Easy, Medium, and Hard levels.}
\label{fig:wider_res}
\end{figure*}

\begin{figure*}
\centering
\subfigure[AFW] {
\includegraphics[height=3.9cm]{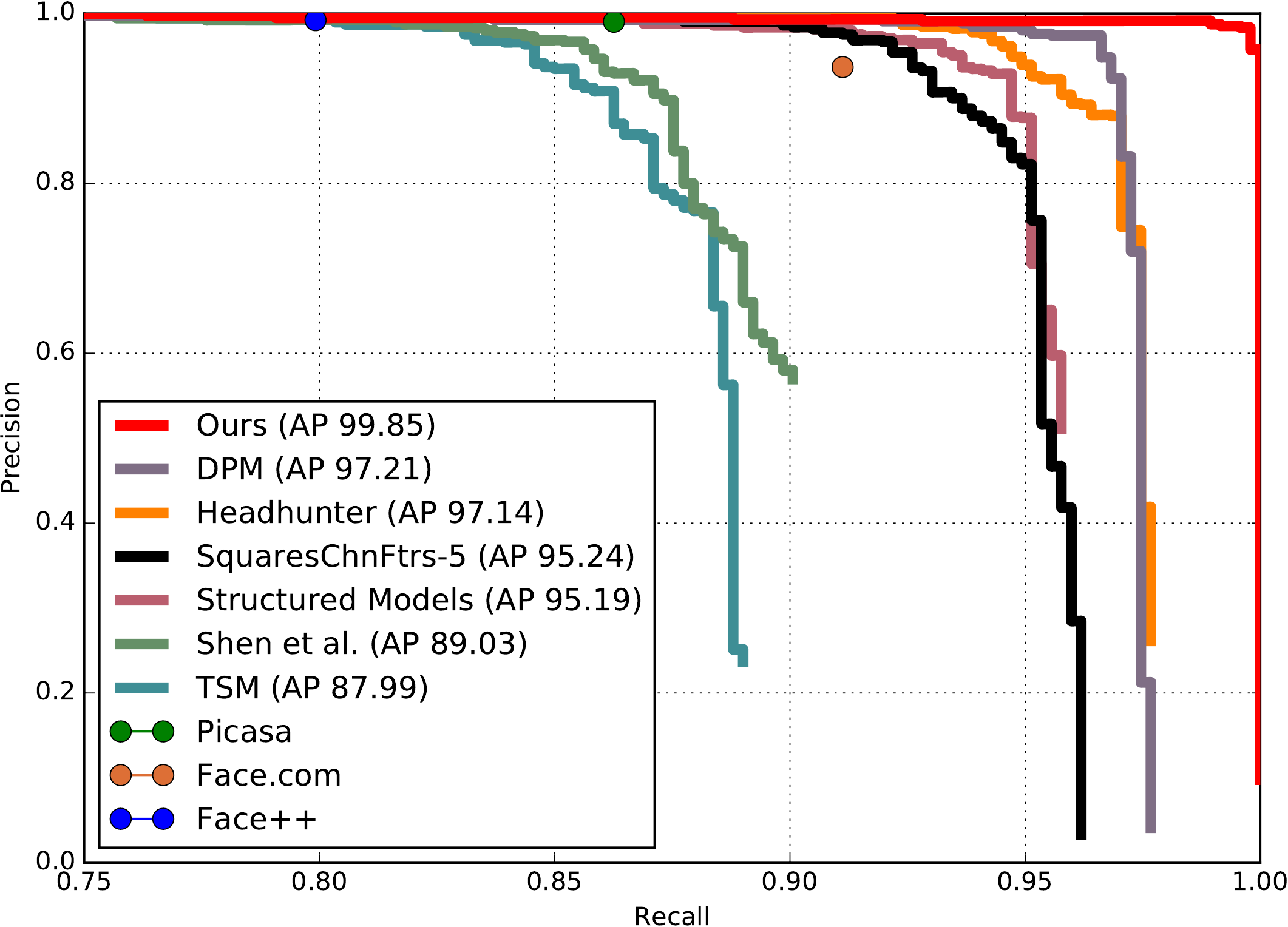}
\label{fig:afw_res}
}
\subfigure[PASCAL Faces] {
\includegraphics[height=3.9cm]{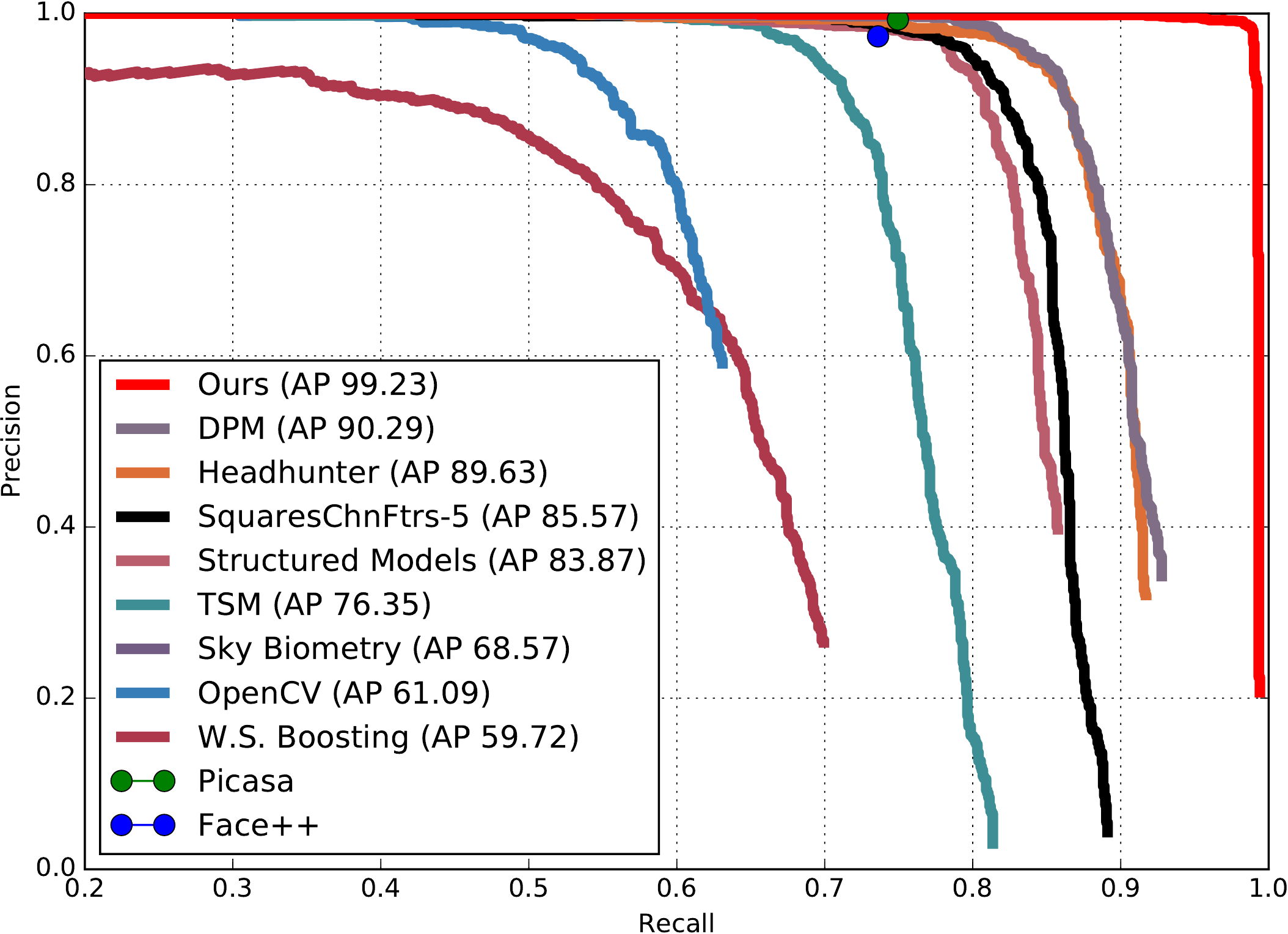}
\label{fig:pascal_res}
}
\subfigure[FDDB] {
\includegraphics[height=3.9cm]{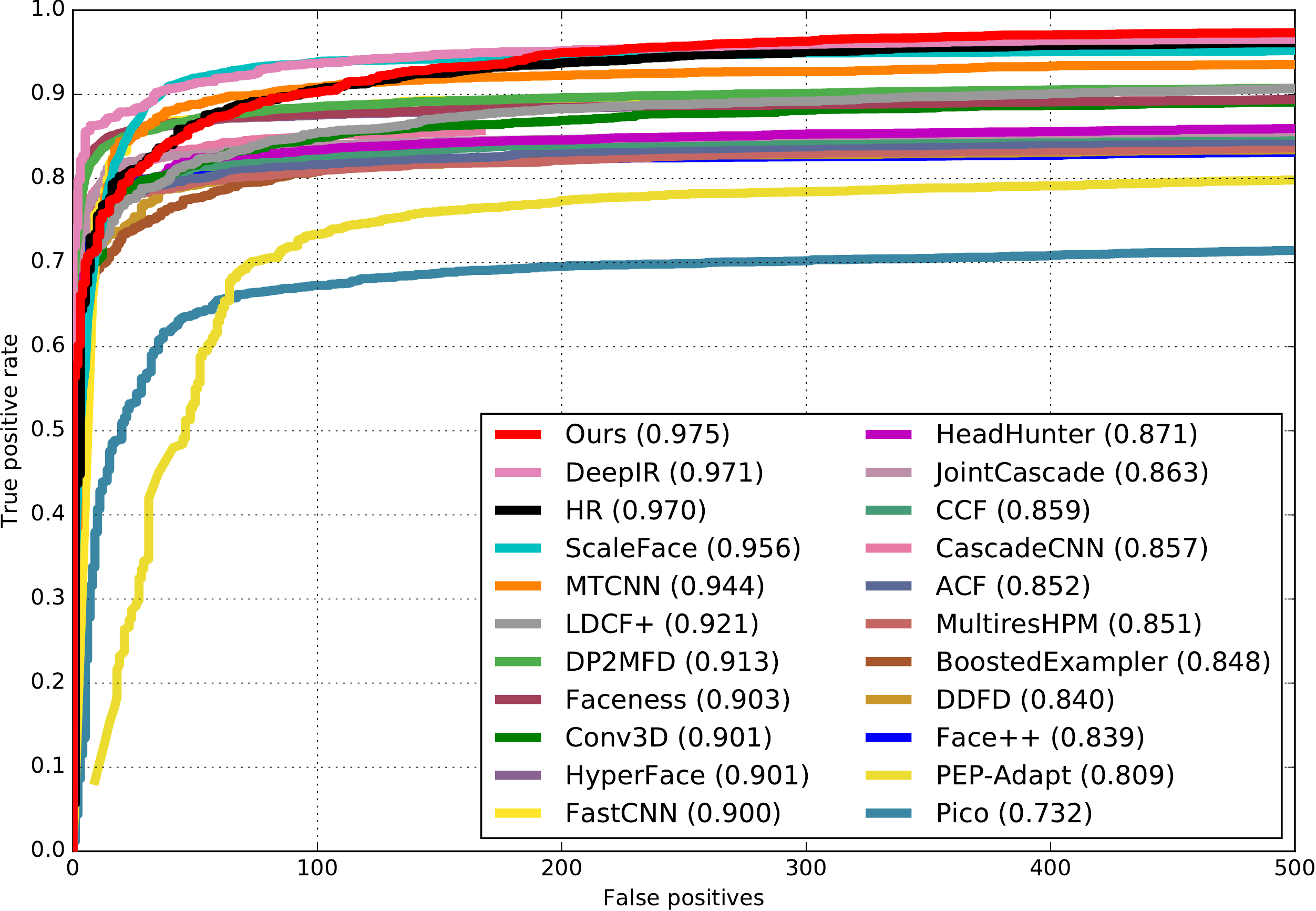}
\label{fig:fddb_res}
}
\caption{Comparison with popular state-of-the-art methods on the AFW, PASCAL Faces, and FDDB datasets}
\end{figure*}

We evaluate our proposed method on the common face detection benchmarks, including Wider Face \cite{yang2016wider}, Annotated Faces in the Wild (AFW) \cite{afw}, PASCAL Faces \cite{faceevaluation15}, and Face Detection Dataset and Benchmark (FDDB) \cite{fddbTech}. Our face detector is trained only using Wider Face training set and is tested on those benchmarks \textit{without} further finetuning. We demonstrate consistent state-of-the-art performance across all the datasets. Qualitative results are illustrated in the supplementary materials.

\textbf{Wider Face dataset}
We report the performance of our face detection system on the Wider Face testing set with 16,097 images. Detection results are sent to the database server for receiving the precision-recall curves. Figure \ref{fig:wider_res} illustrates the precision-recall curves along with AP scores. It's clear that our detector consistently achieves best performance on all face cases against recent published face detection methods: SFD \cite{sfd}, SSH \cite{najibi2017ssh}, ScaleFace \cite{yang2017scaleface}, HR \cite{hu2016tiny}, CMS-RCNN \cite{cms-rcnn}, Multitask Cascade CNN \cite{mtcnn}, LDCF+ \cite{ldcf} and Faceness \cite{faceness}. The results demonstrate that our proposed components further promote the solution for finding small and hard faces.

\textbf{AFW dataset}
This dataset consists of 205 images with 473 annotated faces. During testing images are kept at the original scale. Our method is compared with popular face detection algorithms \cite{mathias2014face, yan2014struct, shen2013detecting,zhu2012face} as well as commercial face detection systems including Picasa, Face.com and Face++. Figure \ref{fig:afw_res} show that our detector outperforms others by a considerable margin.

\textbf{PASCAL Faces dataset}
This dataset is a subset of the PASCAL VOC testing set. It has 851 images with 1,335 annotated faces. Our detector is tested on the original scale of image and compared against popular methods \cite{mathias2014face, yan2014struct, zhu2012face} and some commercial face detection systems (Picasa, Face++). The precision-recall curves in Figure \ref{fig:pascal_res} demonstrate the superiority of our method.

\textbf{FDDB dataset} This dataset has 2,845 images with 5,171 annotated faces. In stead of rectangle boxes, faces in FDDB are represented by ellipses. So we learn a regressor to transform rectangle boxes predicted by our detector to ellipses. Again, we test on the original images without rescaling. We compare our approach with other state-of-the-art methods \cite{deepir,hu2016tiny,yang2017scaleface,mtcnn,ldcf,ranjan2015dp2mfd,faceness,conv3d,ranjan2016hyperface,triantafyllidou2016fast,mathias2014face,chen2014jointcascade,yang2015ccf,li2015cascadecnn,yang2014acf-multiscale,ghiasi2015multireshpm,li2014exemplar,farfade2015ddfd,zhou2013extensive,li2013pep-adapt,markuvs2013pico} which \textit{don't} add additional self-labeled annotations. As shown in Figure \ref{fig:fddb_res}, our detector achieves high recall rate and best AP score.

\subsection{Runtime Speed}
\label{exp:runningtime}
\begin{figure}
\centering
\includegraphics[width=0.7\columnwidth]{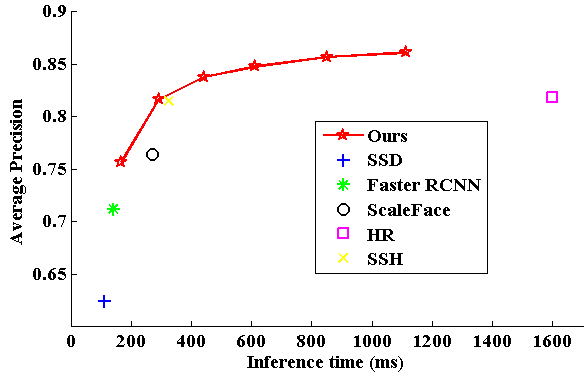}
\caption{Speed (ms) vs. accuracy (AP) on Wider Face Hard set.}
\label{fig:runtime}
\end{figure}
This section reports the runtime speed of our detector on the Wider Face validation set. We vary the image size to get a set of AP scores with corresponded average inference times as shown in Figure \ref{fig:runtime}. Our detector forms an upper envelope of other detectors \cite{liu2016ssd,ren2015faster,yang2017scaleface,hu2016tiny,najibi2017ssh}. Some numbers are acquired from \cite{yang2017scaleface}. All detectors are evaluated using a single NVIDIA Titan X GPU with batch size 1.


\section{Conclusion}
This work identified low face-anchor overlap as the major reason hindering anchor-based detectors to detect tiny faces. We proposed the new EMO score to characterize anchors' capability of getting high overlaps with faces, providing an in-depth analysis of the anchor matching mechanism. This inspired us to come up with several simple but effective strategies of a new anchor design for higher face IoU scores. Consequently, our method outperforms the baseline anchor-based detector by a considerable margin and achieves the state-of-the-art performance on the challenging Wider Face, AFW, PASCAL Faces and FDDB datasets.

{\small
\bibliographystyle{ieee}
\bibliography{egbib}
}

\newpage
\title{Supplementary Materials}
\maketitle
\beginsupplement

\section{Recall Rate Comparison}
\begin{figure}
\centering
\includegraphics[width=\columnwidth]{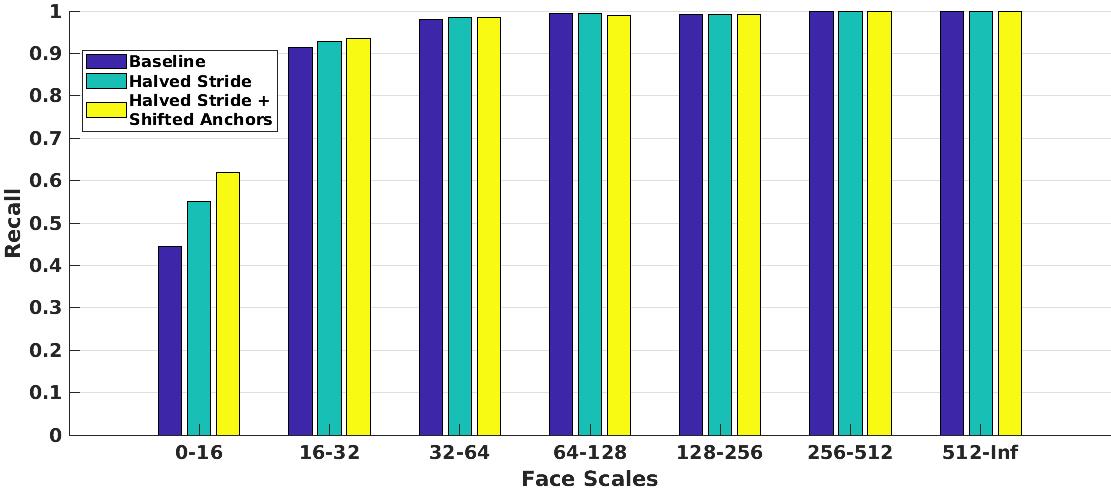}
\caption{The recall rates of the baseline anchor-based detector, and our detector with different components of the new anchor design. Both the proposed anchor stride reduction strategy and extra shifted anchors are effective to help finding more tiny faces.}
\label{fig:recall_cmp}
\end{figure}

This section demonstrates the effect of our new anchor design on the recall rate of anchor-based detectors. As shown in Figure 7 in the main paper, with the new anchor design the average IoU is improved by a large margin for small faces. Here we again observe the positive correlation between the average IoUs and the recall rates in Figure \ref{fig:recall_cmp}. Specifically, both anchor stride reduction from enlarged feature map and extra shifted anchors help to recall more tiny faces. This means it is crucial to have anchors close to faces so that they are easily regressed to face boxes. 

\section{Dilated Convolution}
In the main paper, we propose three network architectures to reduce the anchor stride with enlarged feature maps, i.e. bilinear upsampling, bilinear upsampling with skip connection, and dilated convolution. Experimental results show that dilated convolution extract more robust features than other two architectures, yielding the highest AP scores. In this section we discuss why dilated convolution has such merits.

\begin{figure}
\centering
\includegraphics[width=0.7\columnwidth]{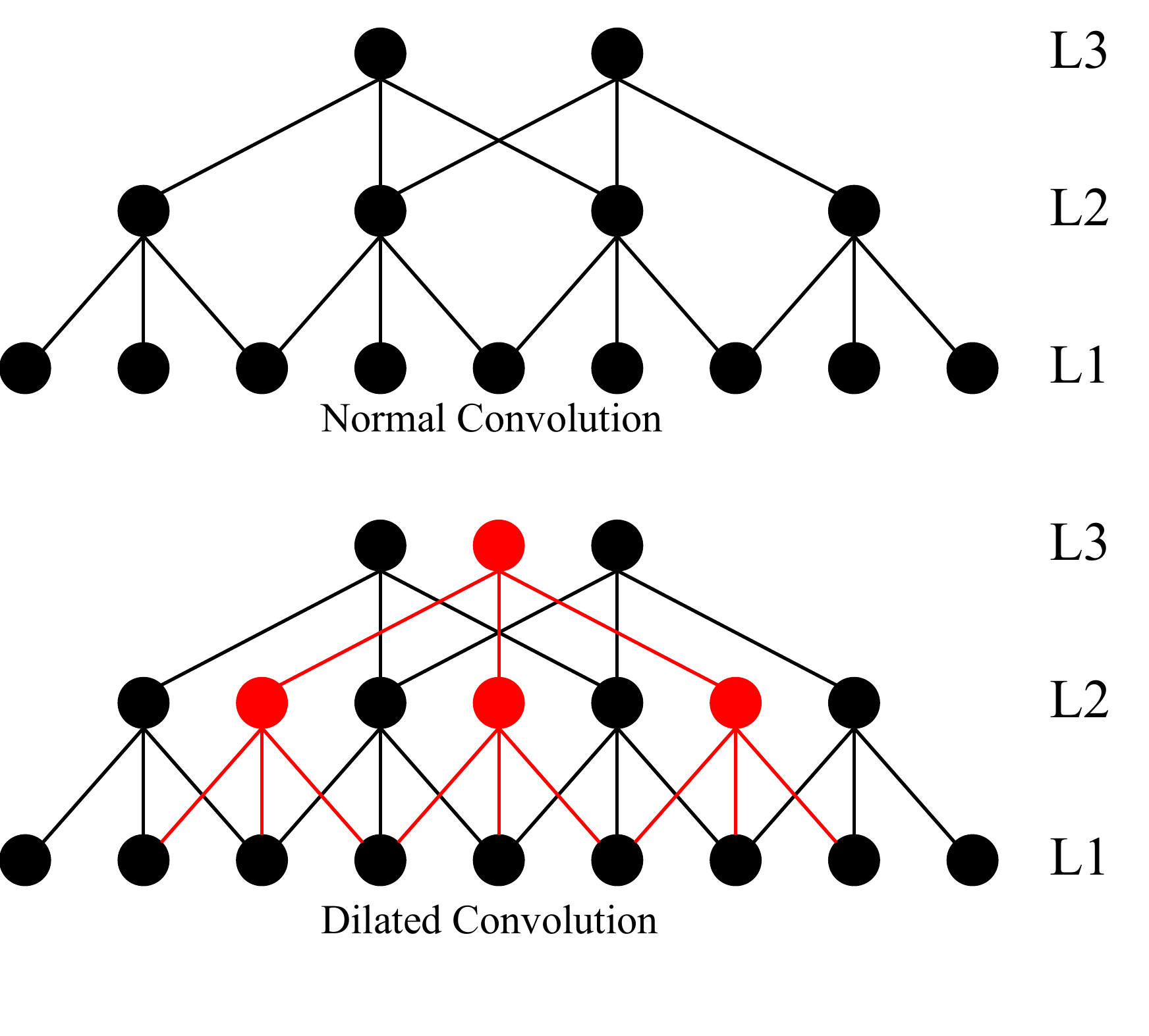}
\caption{A toy example of normal convolution layers (top) and dilated convolution layers (bottom). Difference is marked as red.}
\label{fig:dilation}
\end{figure}

Dilated convolution \cite{dilation} (also known as ``\textit{Algorithme {\`a} trous}'') is widely used in semantic segmentation. Figure \ref{fig:dilation} presents a toy example of comparison between normal convolution layers (top) and dilated convolution layers (bottom). For simplicity we only consider $3 \times 3$ filters and visualize as 1D case. The dots represent the layer units and the lines between dots are connections associated with filter weights. The top part of Figure \ref{fig:dilation} shows a stride-2 convolution followed by a stride-1 convolution without dilation. The bottom part first removes stride-2 operation and then dilates the filter. The difference compared to the normal convolution is labeled in red. It's clear that dilated convolution is actually filling the holes between units.

From the hole filling perspective, we can understand bilinear upsampling as filling the holes with interpolated values, which provides limited new information. And bilinear upsampling with skip connection fills the holes with interpolated values plus features from shallower layer, which are not as discriminative as deep layers. On the other hand, dilated convolution fills the holes layer by layer. The values in the deep layer come from several non-linear transformations of the values in the shallow layer, leading to more discriminative power.

\begin{figure*}
\centering
\subfigure[Easy] {
\includegraphics[width=0.7\columnwidth]{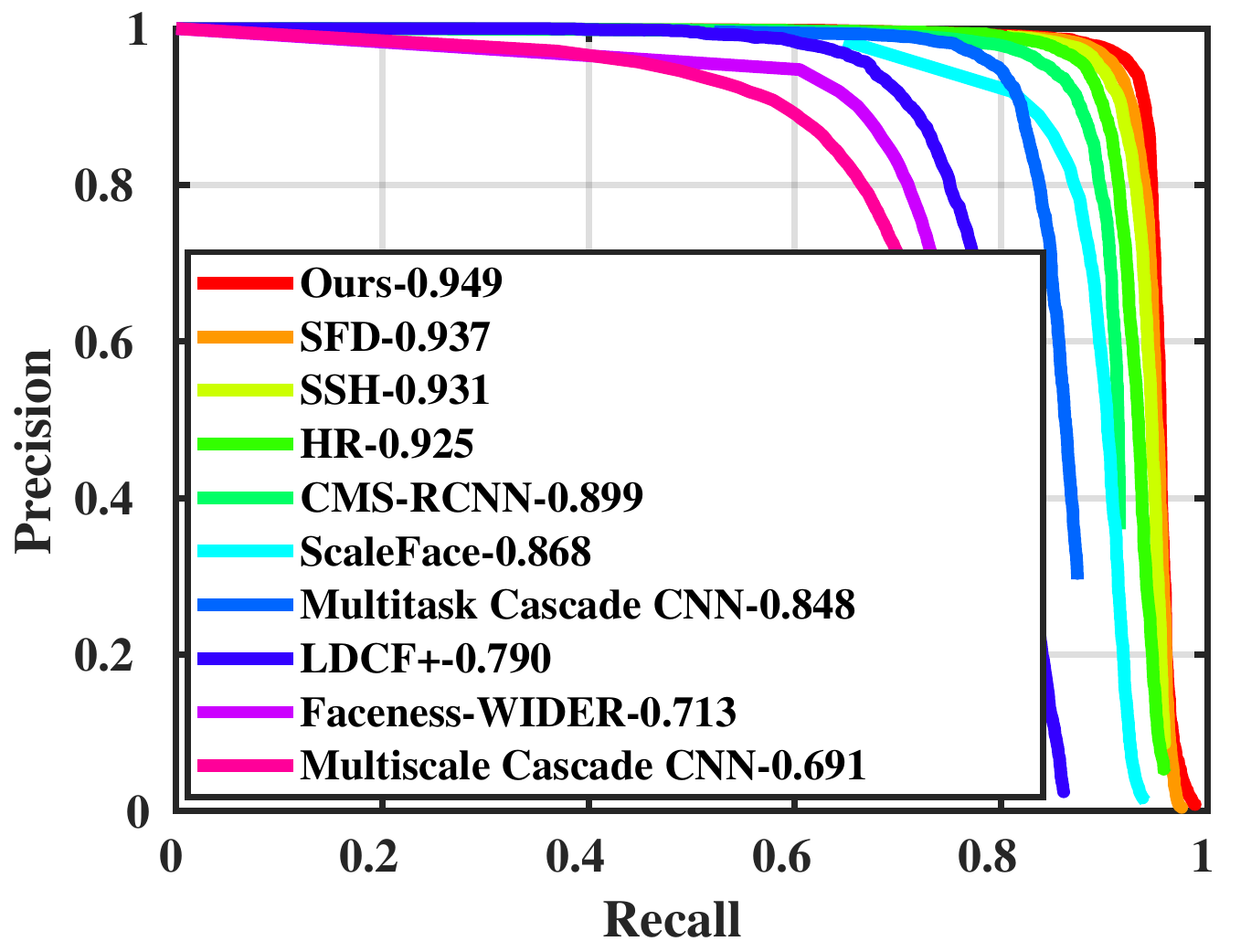}
}
\hspace{-5mm}
\subfigure[Medium] {
\includegraphics[width=0.7\columnwidth]{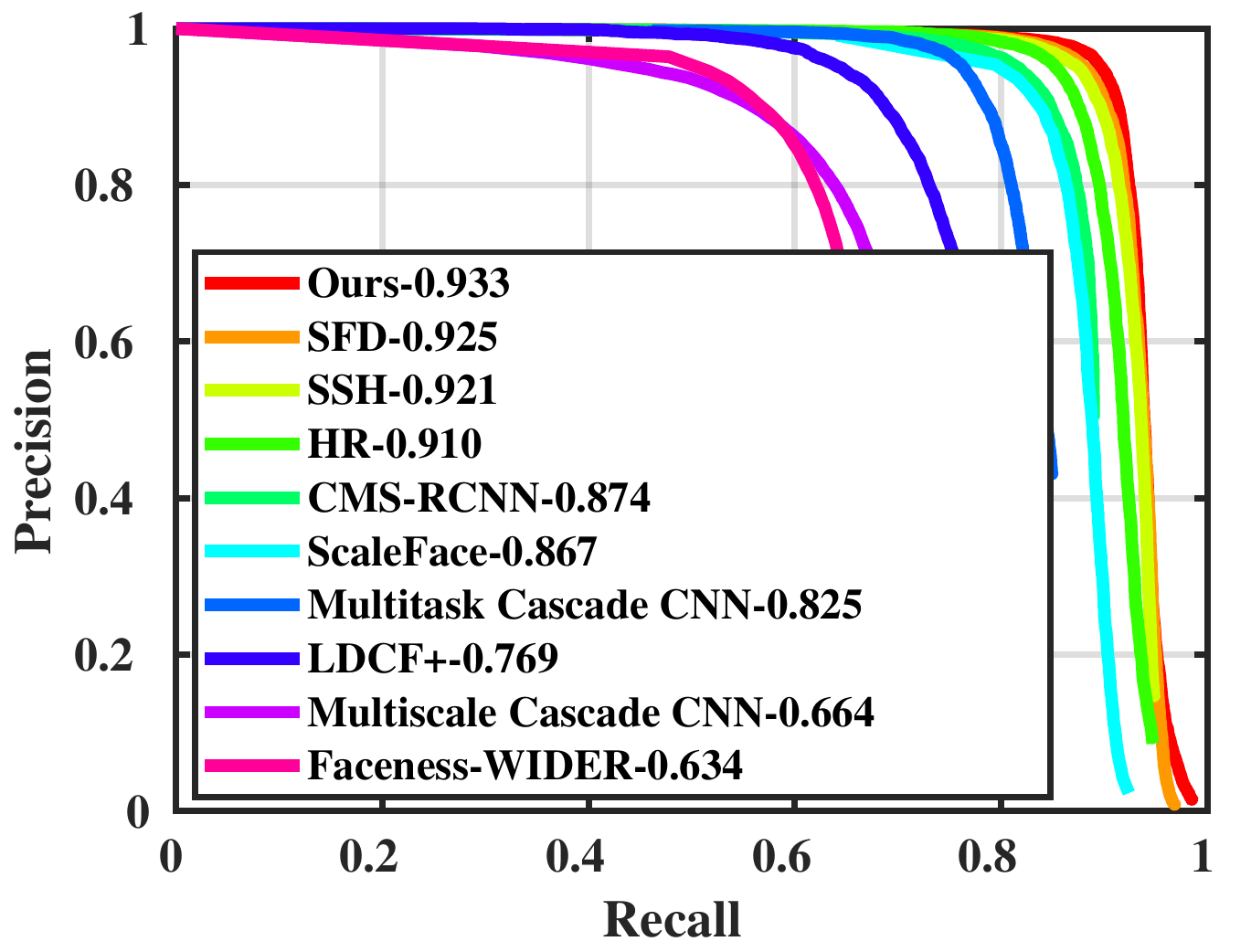}
}
\hspace{-5mm}
\subfigure[Hard] {
\includegraphics[width=0.7\columnwidth]{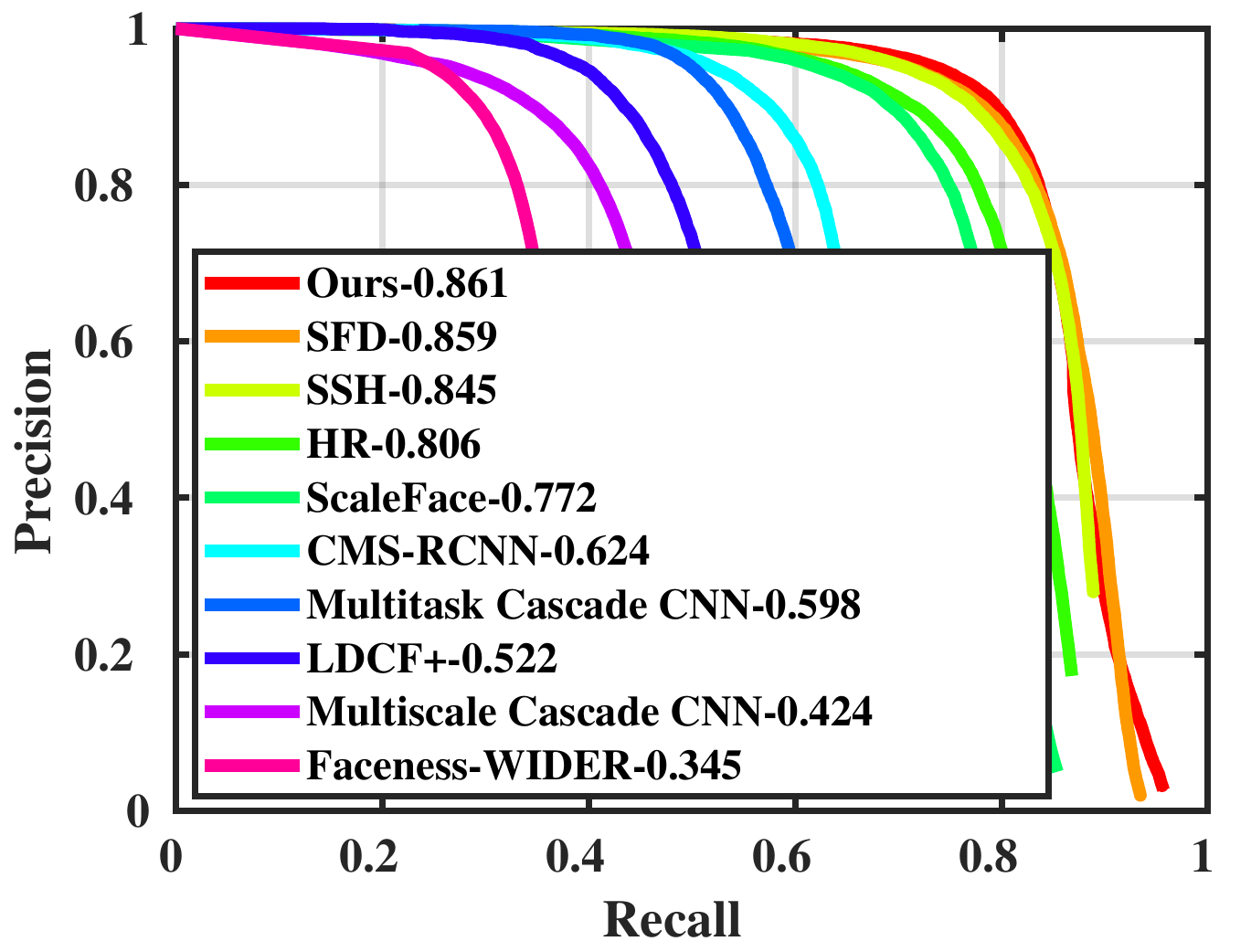}
}
\caption{Precision and recall curves on Wider Face validation set divided into Easy, Medium, and Hard levels.}
\label{fig:pr_wider}
\end{figure*}

\section{More Precision-Recall Curves}
Figure \ref{fig:pr_wider} shows precision-recall curves and AP scores of our detector on the Wider Face \textit{validation set}. Based on our newly proposed anchor design, our detector consistently achieves the state-of-the-art performance across Easy, Medium, and Hard set compared with other recent methods \cite{sfd, najibi2017ssh, hu2016tiny, cms-rcnn, yang2017scaleface, mtcnn, ldcf, faceness, yang2016wider}.

\section{Qualitative Results}
\begin{figure*}
\centering
\includegraphics[width=2\columnwidth]{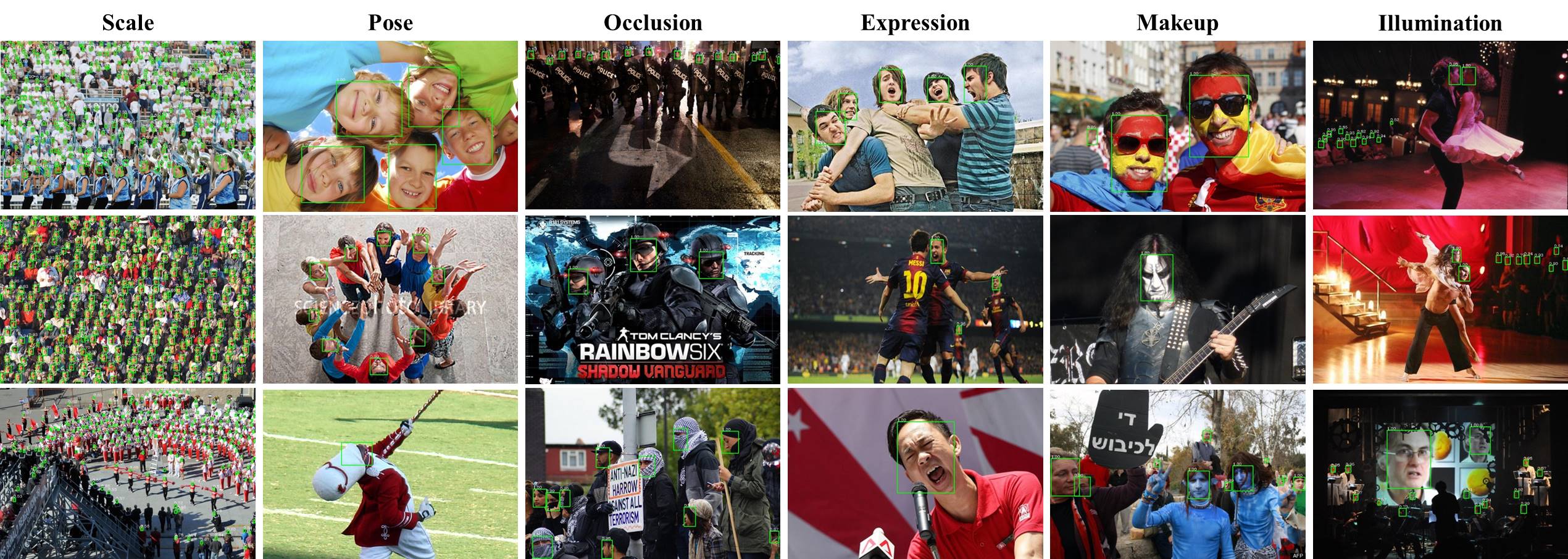}
\caption{Qualitative results on the Wider Face dataset under various challenging conditions, i.e. scale, pose, occlusion, expression, makeup and illumination. Best viewed in color.}
\label{fig:wider}
\end{figure*}

\begin{figure*}
\centering
\includegraphics[width=2\columnwidth]{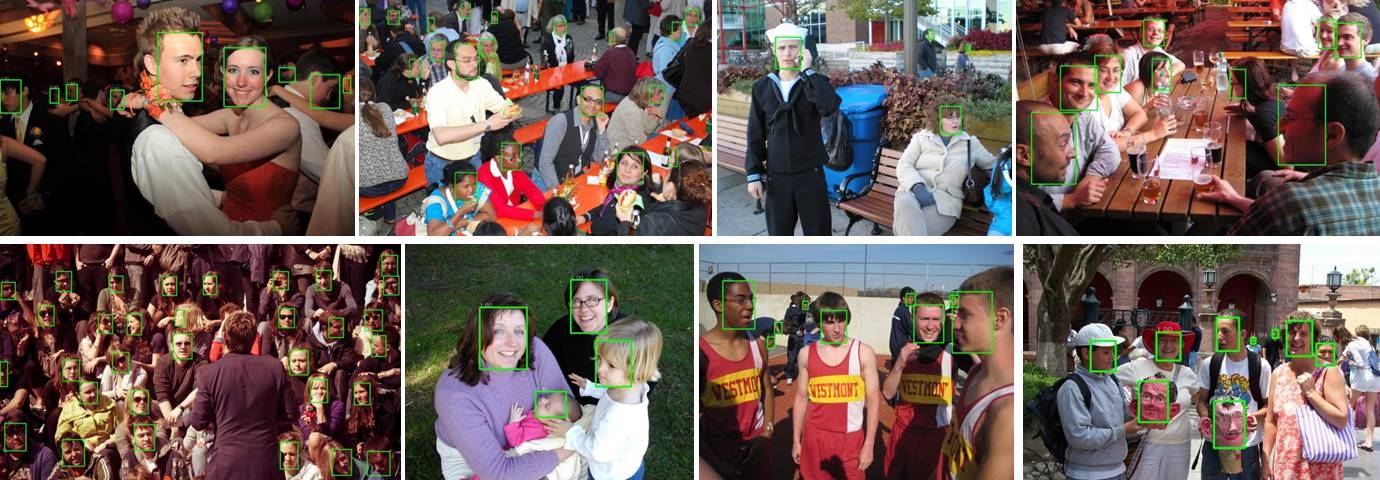}
\caption{Qualitative results on the AFW dataset. Our method can detect challenging faces with high confidence. Best viewed in color.}
\label{fig:afw}
\end{figure*}

\begin{figure*}
\centering
\includegraphics[width=2\columnwidth]{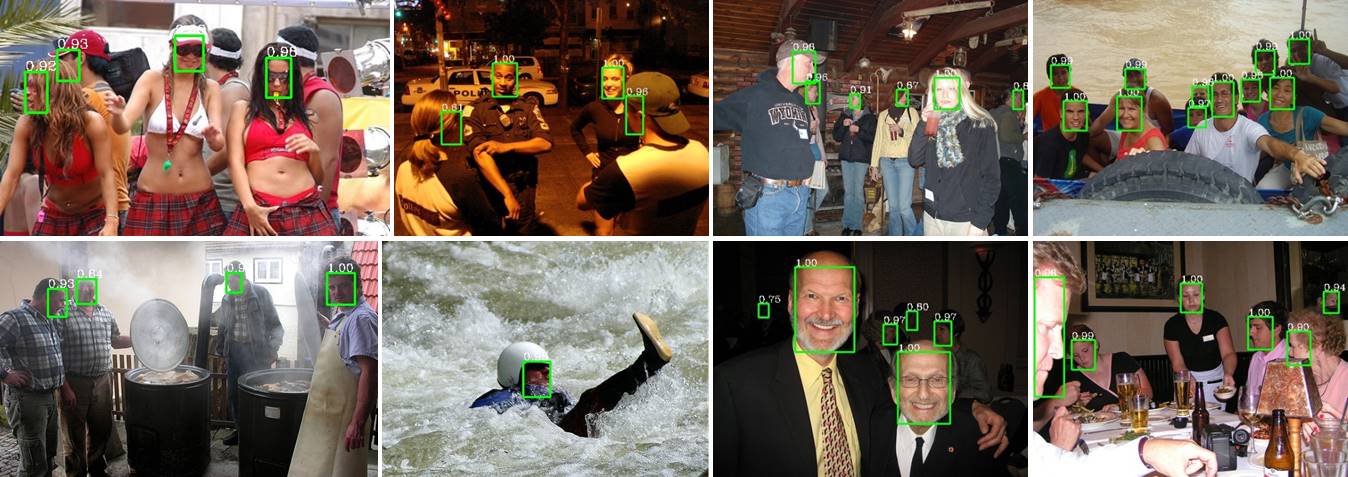}
\caption{Qualitative results on the PASCAL Faces dataset. We can detect challenging faces with high confidence. Best viewed in color.}
\label{fig:pascal}
\end{figure*}

\begin{figure*}
\centering
\includegraphics[width=2\columnwidth]{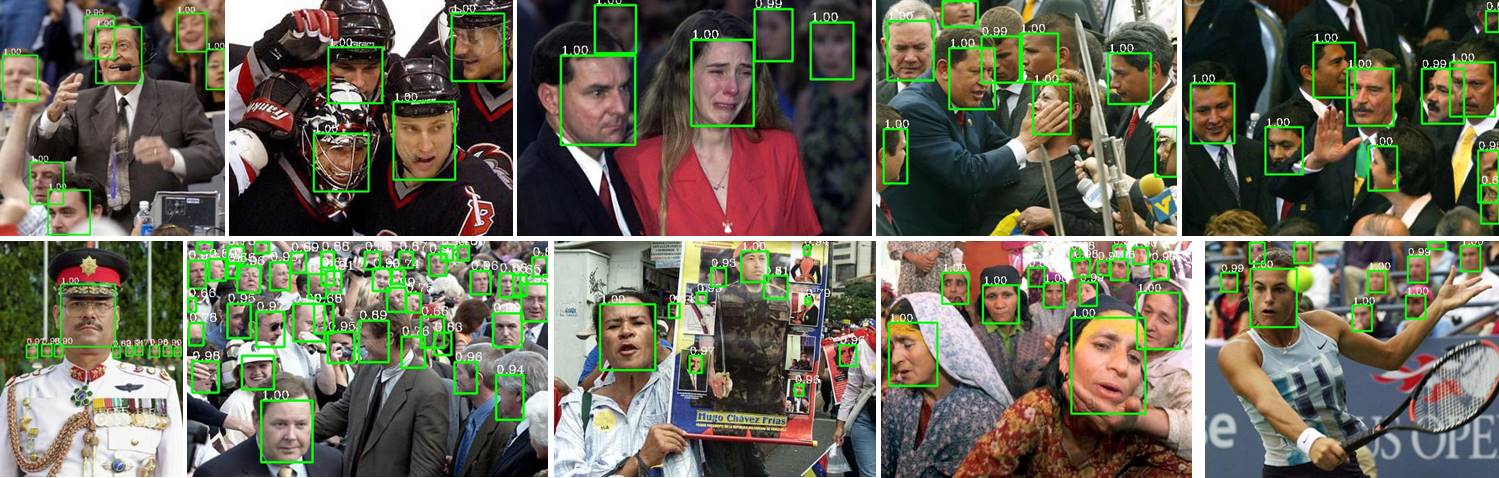}
\caption{Qualitative results on the FDDB dataset. Best viewed in color.}
\label{fig:fddb}
\end{figure*}

\begin{figure*}
\centering
\includegraphics[width=2\columnwidth]{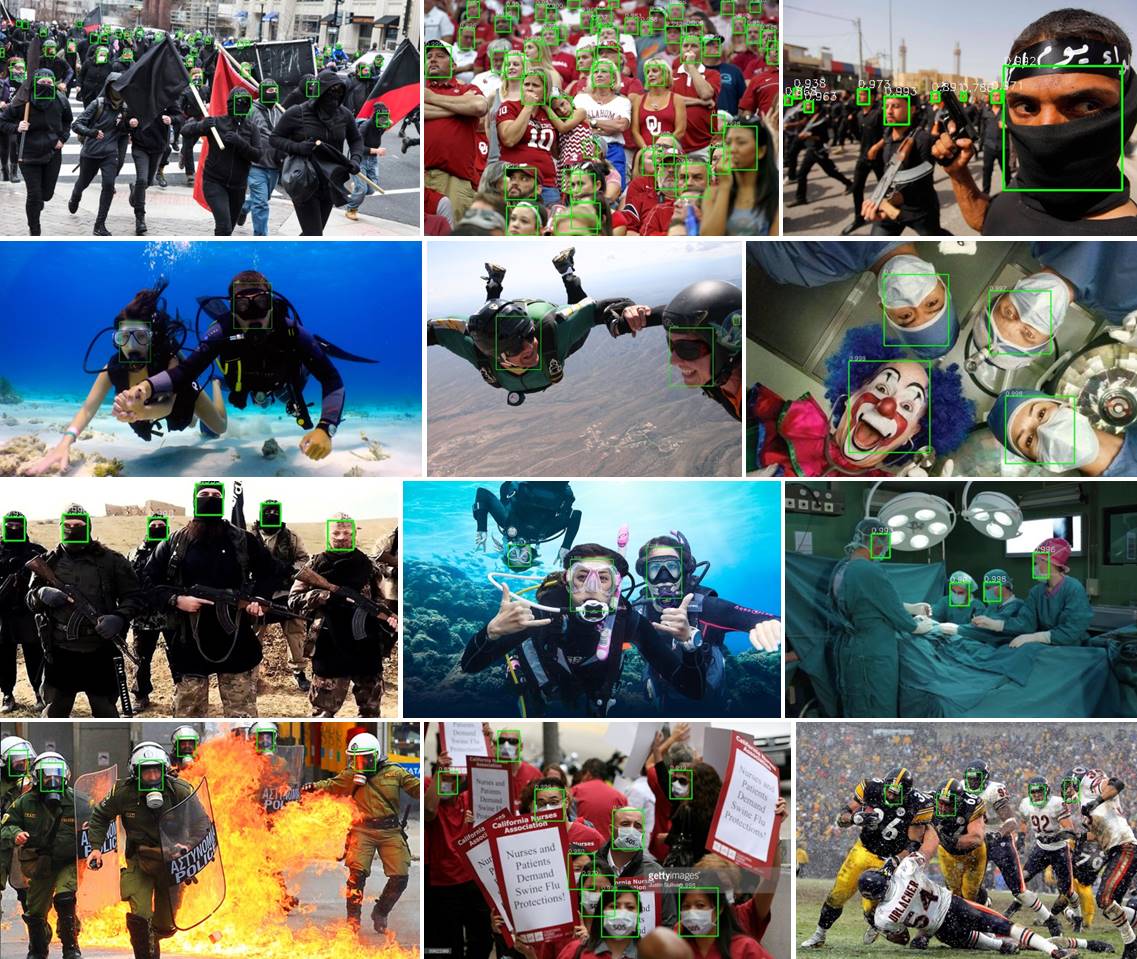}
\caption{Qualitative results on some interesting images from the Internet. Best viewed in color.}
\label{fig:others}
\end{figure*}

This section demonstrates some qualitative results on evaluation benchmarks including Wider Face (Figure \ref{fig:wider}), AFW (Figure \ref{fig:afw}), PASCAL Faces (Figure \ref{fig:pascal}), FDDB (Figure \ref{fig:fddb}), as well as on several interesting images from the Internet (Figure \ref{fig:others}). We annotate the detected faces with green boxes and show the confidence scores on the top of boxes. Please zoom in to see the detections of tiny faces.

\end{document}